\pdfoutput=1

\documentclass[11pt]{article}
\usepackage[final]{acl}

\usepackage{times}
\usepackage{latexsym}

\usepackage[T1]{fontenc}

\usepackage[utf8]{inputenc}

\usepackage{microtype}

\usepackage{inconsolata}

\usepackage{graphicx}

\usepackage{hyperref}
\usepackage{url}
\usepackage{booktabs}
\usepackage{lineno}
\usepackage{amsfonts}
\usepackage{nicefrac}
\usepackage{float}
\usepackage{amsmath}
\usepackage{subcaption}
\usepackage{array}
\usepackage{adjustbox}
\usepackage{multirow}
\usepackage{tabularx}
\usepackage{caption}
\usepackage{placeins}
\usepackage{colortbl}
\usepackage{siunitx}
\usepackage{makecell}
\usepackage{tikz}
\usepackage{eqparbox}
\usepackage{mathtools}
\usepackage{annotate-equations}
\usepackage{textcomp, gensymb}
\usepackage{xcolor} 
\usepackage{soul}
\usepackage{wrapfig}
\usepackage{ntheorem}
\usepackage{enumitem}

\definecolor{lightgreen}{RGB}{204, 255, 204}
\definecolor{lightred}{RGB}{255, 204, 204}
\newcommand{\ckconversion}[1]{\cellcolor{lightgreen} #1}
\newcommand{\pk}[1]{{\sethlcolor{cyan!15}\hl{#1}}}
\newcommand{\ck}[1]{{\sethlcolor{lightred}\hl{#1}}}

\title{Instructions for *ACL Proceedings}

\author{
    \textbf{Zineddine Tighidet\textsuperscript{1, 2}}, 
    \textbf{Andrea Mogini\textsuperscript{1}}, 
    \textbf{Hedi Ben-younes \textsuperscript{1}}, 
    \textbf{Jiali Mei\textsuperscript{1}},
    \\
    \textbf{Patrick Gallinari\textsuperscript{2, 3}}, 
    \textbf{Benjamin Piwowarski\textsuperscript{2}}
    \\
    \\
    \textsuperscript{1}BNP Paribas, Paris, France
    \\
    \textsuperscript{2}Sorbonne Université, CNRS, ISIR, F-75005 Paris, France
    \\
    \textsuperscript{3}Criteo AI Lab, Paris, France
    \\
    \normalsize{
        Correspondence: \textit{zineddine.tighidet@}\{\textit{bnpparibas.com, sorbonne-universite.fr}\}
    }
}

\definecolor{darkblue}{rgb}{0, 0, 0.5}
\definecolor{lowq}{RGB}{198,239,206} 
\definecolor{highq}{RGB}{255,199,206} 
\definecolor{purple}{HTML}{735b8a}
\definecolor{blue_drawio}{HTML}{6C8EBF}
\definecolor{green_drawio}{HTML}{82B366}
\definecolor{dark_green_drawio}{HTML}{557543}
\definecolor{lightblue}{RGB}{175, 210, 230}
\definecolor{lightorange}{RGB}{255, 210, 170}
\definecolor{lightgreen}{RGB}{190, 230, 190}
\definecolor{lightred}{RGB}{255, 204, 204}
\sethlcolor{lightred}

\title{Context Copying Modulation: The Role of Entropy Neurons in Managing Parametric and Contextual Knowledge Conflicts}

\begin{document}

\maketitle

\begin{abstract}
The behavior of Large Language Models (LLMs) when facing contextual information that conflicts with their internal parametric knowledge is inconsistent, with no generally accepted explanation for the expected outcome distribution. Recent work has identified in autoregressive transformer models a class of neurons -- called \textit{entropy neurons} -- that produce a significant effect on the model output entropy while having an overall moderate impact on the ranking of the predicted tokens. In this paper, we investigate the preliminary claim that these neurons are involved in inhibiting context copying behavior in transformers by looking at their role in resolving conflicts between contextual and parametric information. We show that \textit{entropy neurons} are responsible for suppressing context copying across a range of LLMs, and that ablating them leads to a significant change in the generation process. These results enhance our understanding of the internal dynamics of LLMs when handling conflicting information.\footnote{We make our code and data publicly available at: \href{https://github.com/Zineddine-Tighidet/Context-Copying-Modulation}{https://github.com/Zineddine-Tighidet/Context-Copying-Modulation}}
\end{abstract}

\begin{figure}[t]
    \centering
    \includegraphics[width=1\columnwidth]{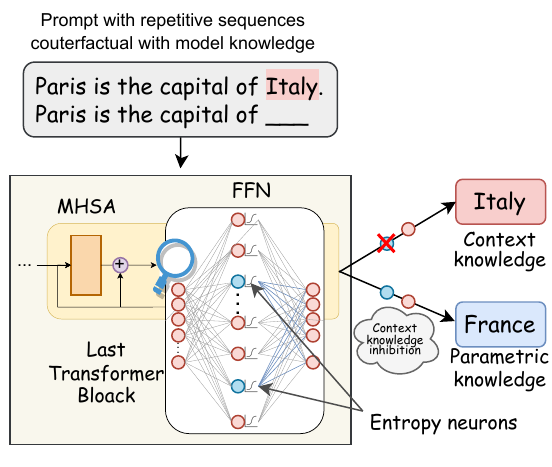}
    \caption{Schema illustrating how entropy neurons influence LLMs' decision-making between the provided contextual knowledge (CK) and the learned parametric knowledge (PK). By presenting the model with repetitive prompts contradicting its PK, we test whether it relies on PK or CK. Ablating entropy neurons reveals their causal role: they inhibit the use of CK (e.g., \ck{Italy}) in favor of PK (e.g., \pk{France}).}
    \label{fig:schema_methodo}
\end{figure}

\section{Introduction}

Large Language Models (LLMs) exhibit remarkable proficiency in representing, memorizing, and retrieving vast amounts of information. However, they often struggle when discrepancies arise between their learned \textbf{parametric knowledge (PK)} and the \textbf{contextual knowledge (CK)} provided at inference \citep{xie2024adaptive, jin2024cuttingheadendsconflict, xu2024knowledgeconflictsllmssurvey}. These conflicts can lead to unpredictable model behavior, which poses a significant challenge in real-world applications \citep{Ji_2023}.

\begin{table*}[!t]
\centering
\footnotesize
\begin{tabular}{p{0.62\textwidth}p{0.13\textwidth}p{0.13\textwidth}}
\toprule
\textbf{Input Prompt} & \textbf{Before (PK)} & \textbf{After (CK)} \\
\midrule
Kentucky's official language is \ck{Japanese}. Kentucky's official language is & \pk{English} & \ck{Japanese} \\
\addlinespace
Antonio Moreno communicates in \ck{English}. Antonio Moreno communicates in & \pk{Spanish} & \ck{English} \\
\addlinespace
Mac OS X Panther is a product released by \ck{Google}. Mac OS X Panther is a product released by & \pk{Apple} & \ck{Google} \\
\bottomrule
\end{tabular}
\caption{Examples where Phi-1.5 switched from using \textbf{Parametric Knowledge (PK)} to \textbf{Contextual Knowledge (CK)} after ablating entropy neurons.}
\label{tab:pk_to_ck_switch_examples}
\end{table*}

\textcolor{black}{Although various strategies have been proposed to mitigate this unpredictable behavior \citep{NEURIPS2024_08a9e28c}}, the mechanisms that govern how LLMs prioritize and integrate different sources of knowledge are poorly understood. \textcolor{black}{Understanding these mechanisms is crucial, as the resolution of PK–CK conflicts directly impacts context-intensive tasks such as retrieval augmented generation (RAG) and other applications where accuracy depends on balancing internal knowledge with external context. Without a clear regulation process, models may either ignore reliable contextual cues or override their own parametric knowledge inappropriately.}

We investigate the preliminary claim that the recently discovered \textit{entropy neurons} \citep{katz-belinkov-2023-visit, gurnee2024universal} are involved in inhibiting context copying behavior \citep{stolfo2024confidenceregulationneuronslanguage} by looking at their role in resolving conflicts between CK and PK. \textcolor{black}{These neurons are known to regulate model entropy while having an overall moderate impact on the ranking of the predicted tokens.}
\textcolor{black}{By investigating entropy neurons, we aim to uncover the mechanisms governing this balance and provide insights into how LLMs integrate different knowledge sources in practice.}

\textcolor{black}{Understanding this balance mechanism is critical for developing more reliable and grounded language models. By elucidating how entropy neurons mediate these conflicts, we establish empirical grounds for targeted interventions that could enforce more consistent knowledge integration. This mechanistic understanding enables the development of safer models with reduced propensity for extrinsic and intrinsic hallucinations \cite{Ji_2023, bang2025hallulensllmhallucinationbenchmark}.}

We make the following key findings and contributions:

\begin{itemize}
    \item Entropy neurons, although representing less than 2\textperthousand{} of the feed forward network neurons in the last transformer layer, play a significant role in determining the knowledge source to use. More specifically, they inhibit the natural LLM's behavior of repeating the sequences in the context, i.e. induction \citep{olsson2022incontextlearninginductionheads}.
    \item We identify the presence of entropy neurons in a range of models, from 1 billion to 8 billion parameters, including Pythia-1.4B, Phi-1.5, Mistral-7B-v0.1, and Llama-3-8B\footnote{In the main paper we show results for Phi-1.5, we provide the results for other models in the Appendix.} and give some insights on their characteristics.
\end{itemize}


\section{Related Work}

The understanding of the mechanisms and knowledge localization within transformers has advanced through various studies. One line of research has focused on the PK-based outputs, particularly in factual settings \citep{geva2021transformer, heinzerling-inui-2021-language, alkhamissi2022review, meng2023locating, geva2023dissecting}. These studies hypothesized that LLMs store parametric information within the Feed Forward Network (FFN) layers, which function as a key-value memory. This stored information is subsequently accessed by the Multi-Head Self-Attention (MHSA) layers. Another body of work has examined CK-based outputs. These studies concluded that the processing of CK, unlike PK, is not localized within the LLM’s parameters \citep{monea2024glitch}. Instead, it is facilitated by a learned mechanism known as induction, which underpins in-context learning and information copying \citep{olsson2022incontextlearninginductionheads}. \textcolor{black}{Despite these advancements, the mechanisms underlying how LLMs regulate the CK usage in a situation of induction are not well understood.}

\section{Background}
\subsection{\textcolor{black}{Feed Forward Network (FFN)}} \textcolor{black}{The structure of the Transformer's FFN is central to our study \cite{NIPS2017_3f5ee243}. Given a hidden state $x \in \mathbb{R}^{d_{model}}$ from the residual stream after the MHSA module, the FFN is defined as:}

\vspace{-0.2cm}
{\small
\begin{equation} \label{eq:mlp}
\textcolor{black}{\mathrm{FFN}(\mathbf{x}) = \sum_{i} w_{\mathrm{out}}^{(i)} \sigma \left( w_{\mathrm{in}}^{(i)} \cdot x + \beta_{\mathrm{in}}^{(i)} \right) + \beta_{\mathrm{out}},
}
\end{equation}}
\noindent where $\mathbf{W}^T_\mathrm{out}, \mathbf{W}_\mathrm{in} \in \mathbb{R}^{d_\mathrm{\text{ffn}} \times d_\mathrm{model}}$ are learned weight matrices, $\boldsymbol{\beta}_\mathrm{in}$ and $\boldsymbol{\beta}_\mathrm{out}$ are learned biases. The function $\sigma$ denotes an element-wise nonlinear activation function, e.g. ReLU \citep{agarap2019deeplearningusingrectified}.

A neuron from the first FFN layer is characterized by 1) an activation value noted $n_i \in \mathbb{R}$ (i.e. the output of the activation function $\sigma$) and 2) an output weight vector $w_{\mathrm{out}}^{(i)} \in \mathbb{R}^{d_{model}}$.
\subsection{Framework and Dataset}
\label{subsection:framework}
We use the knowledge probing framework \cite{tighidet2024probinglanguagemodelsknowledge}, which consists of a dataset of prompts that are built to contradict the internal knowledge (i.e. PK) of a given model. It follows a well-structured format based on repetition, which makes it convenient for PK/CK analysis. A similar framework is proposed by \citet{yu-etal-2023-characterizing} \textcolor{black}{but it consists of prompts in form of questions rather than repetitive sequences which is less convenient to study induction.} \textcolor{black}{We provide characteristics about the dataset in Appendix \ref{sec:data_characteristics}.}

\textcolor{black}{Each prompt $x$ from the knowledge probing dataset $E$} consists of a contextual statement about a subject $s$ (e.g., \textit{"Paris"}), a relation $r$ (e.g., \textit{"capital of"}), and an object $\bar{o}$ that contradicts the model's internal PK (e.g., \textit{"Italy"}). \textcolor{black}{The contextual statement is from the CK that is defined below.} This is followed by a repetitive query about $s$ to trigger the model's induction mechanism. For example:

\begin{align*}
\hspace{-0.5cm}\eqnmark[purple]{context}{\texttt{Paris is the capital of Italy.}}\\
\eqnmark[blue_drawio]{query}{\texttt{Paris is the capital of}}\eqnmarkbox[dark_green_drawio]{attribute}{\text{\rule{1.1cm}{0.1mm}}}
\annotate[yshift=0em]{above, left, label below}{context}{Context Statement}
\annotate[xshift=-1em]{below, left, label above}{query}{Query}
\annotate[yshift=0em]{below, left, label above}{attribute}{Object to predict}
\end{align*}\\
If the model responds according to the context statement, it uses CK (e.g. \textit{"Italy"}). If it responds based on its learned knowledge, it uses PK (e.g. \textit{"France"}). If it outputs neither, the knowledge source is not defined (ND, e.g. \textit{"Spain"}).

\paragraph{Parametric Knowledge (PK).} PK is the information the model learned during training, represented as triplets $(s, r, o)$ where $o$ is the generated object given a query with a subject $s$ and a relation $r$ (e.g., Query: \textit{"Paris is the capital of"} $\rightarrow$ Model answer: \textit{"France"}).

\paragraph{Context Knowledge (CK).} \textcolor{black}{CK is the information that is contradictory to PK.} This involves replacing $o$ with another object $\bar{o}$ that shares the same relation $r$ (e.g., \textit{"Paris is the capital of Italy"}). For each $(s,r)$ couple, three $\bar{o}$ objects are selected, namely those with the lowest probability.
%
%
This selection method ensures the model did not learn the ($s$, $r$, $\bar{o}$) association from its training data.

\paragraph{Not Defined Knowledge (ND).} ND includes all objects not in PK or CK.

\paragraph{Decoding strategy.} \textcolor{black}{Following the knowledge probing framework \cite{tighidet2024probinglanguagemodelsknowledge}, we use a greedy decoding strategy to generate outputs. This deterministic decoding ensures that the results are not influenced by sampling noise (e.g., from temperature or beam search variations).}

\section{Entropy Neurons}
\label{sec:entropy_neurons}
\subsection{Motivation}
\label{sec:entropy_neurons_def}
\citet{gurnee2024universal} and \citet{stolfo2024confidenceregulationneuronslanguage} identified entropy neurons in GPT-2 by considering the 6 neurons with the lowest impact on logits variance using the LogitVar measure, defined in Eq.~\ref{eq:logitvar} \textcolor{black}{and questioned their high weight norm.} \citet{stolfo2024confidenceregulationneuronslanguage} \textcolor{black}{characterize} entropy neurons \textcolor{black}{as those that} write into the \textcolor{black}{effective null space} of the unembedding matrix $\mathbf{W}_\mathrm{U} \in \mathbb{R}^{V\times d_{model}}$, as measured by $\rho$ (Eq.~\ref{eq:rho}).

\paragraph{LogitVar.} This measure quantifies a neuron's direct effect on output logits variance. For a neuron $i$, it is defined as:

{
\small
\begin{equation}
    \label{eq:logitvar}
    \mathrm{LogitVar}(w_{\mathrm{out}}^{(i)}) = \textbf{Var}\left\{ 
        \frac{
           {w}^{(t)}_\mathrm{U} 
           \cdot
           w_{\mathrm{out}}^{(i)}
        }{
            ||{w}^{(t)}_{\mathrm{U}}|| 
            \times
            ||w_{\mathrm{out}}^{(i)}||
        }
        ;
        t \in V
    \right\}
\end{equation}
}
where $V$ is the set of tokens in the vocabulary and $w^{(t)}_\mathrm{U}$ the $t^{\mathrm{th}}$ row of $\mathbf{W}_\mathrm{U}$.

\paragraph{\textcolor{black}{Effective Null Space} Projection ($\rho$).} This measure quantifies how much of a neuron's output aligns with directions that minimally impact the model's final output, forming the \textcolor{black}{effective null space} of the unembedding matrix $\mathbf{W}_\mathrm{U}$, denoted as $\mathbf{V}_\mathrm{0}$. Details on identifying $\mathbf{V}_\mathrm{0}$ are in Appendix \ref{sec:null_space_identification}. For a neuron $i$, it is defined as:

\small
\begin{align}
    \label{eq:rho}
    \rho_i = \frac{||\mathbf{V}_\mathrm{0}^\mathrm{T} w_{\mathrm{out}}^{\mathrm{(i)}}||}{||w_{\mathrm{out}}^{\mathrm{(i)}}||}.
\end{align}
\normalsize
\paragraph{Why are they called "entropy" neurons?}
\textcolor{black}{The term entropy neurons was introduced by \citet{gurnee2024universal}, who observed that these neurons influence the entropy of the model's output distribution while affecting minimally the relative ranking of tokens.
}

\paragraph{Why are they interesting?}
\textcolor{black}{Our interest in these neurons stems from preliminary findings by \citet{stolfo2024confidenceregulationneuronslanguage}, which suggest that induction heads—attention heads associated with context-copying behavior—causally affect entropy neurons. This connection raises the intriguing possibility that entropy neurons may play a role in regulating copy behavior in transformer models.
}

\begin{figure*}[t]
    \begin{subfigure}[b]{0.49\textwidth}
        \centering
        \includegraphics[width=\textwidth]{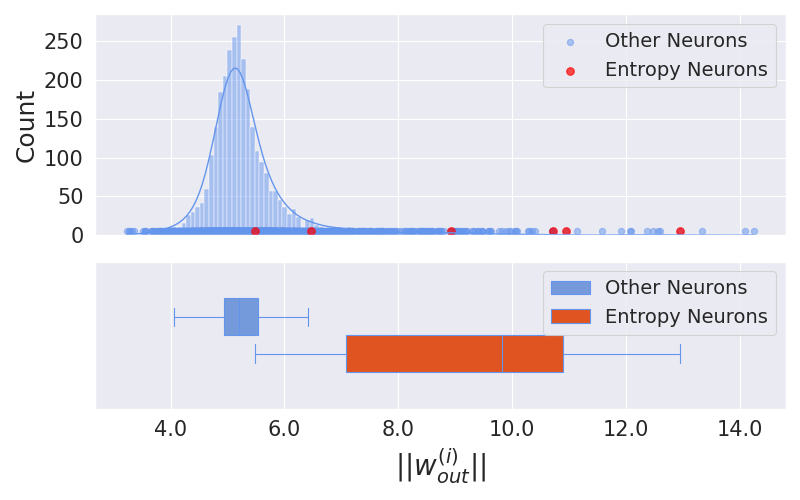}
        \caption{GPT-2}
        \label{fig:weigh_norm_distro_gpt2}
    \end{subfigure}
    \hspace{-0.35cm}
    \begin{subfigure}[b]{0.49\textwidth}
        \centering
        \includegraphics[width=\textwidth]{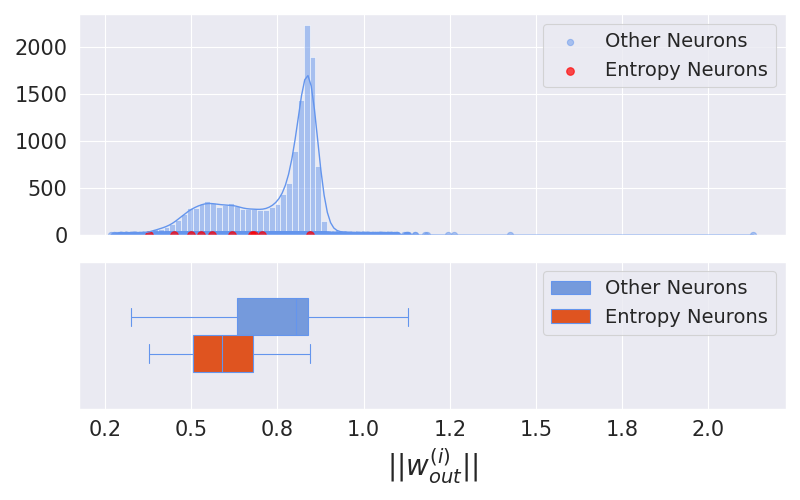}
        \caption{Llama-3-8B}
        \label{fig:weigh_norm_distro_llama3}
    \end{subfigure}
    \caption{Weight norm distribution for entropy neurons vs. other neurons for GPT-2 and Llama-3-8B. Llama-3-8B entropy neurons's, in contrast to GPT-2, exhibit a lower weight norm compared to other neurons.}
    \label{fig:weigh_norm_distro}
\end{figure*}

\begin{figure}[t]
    \includegraphics[width=\columnwidth]{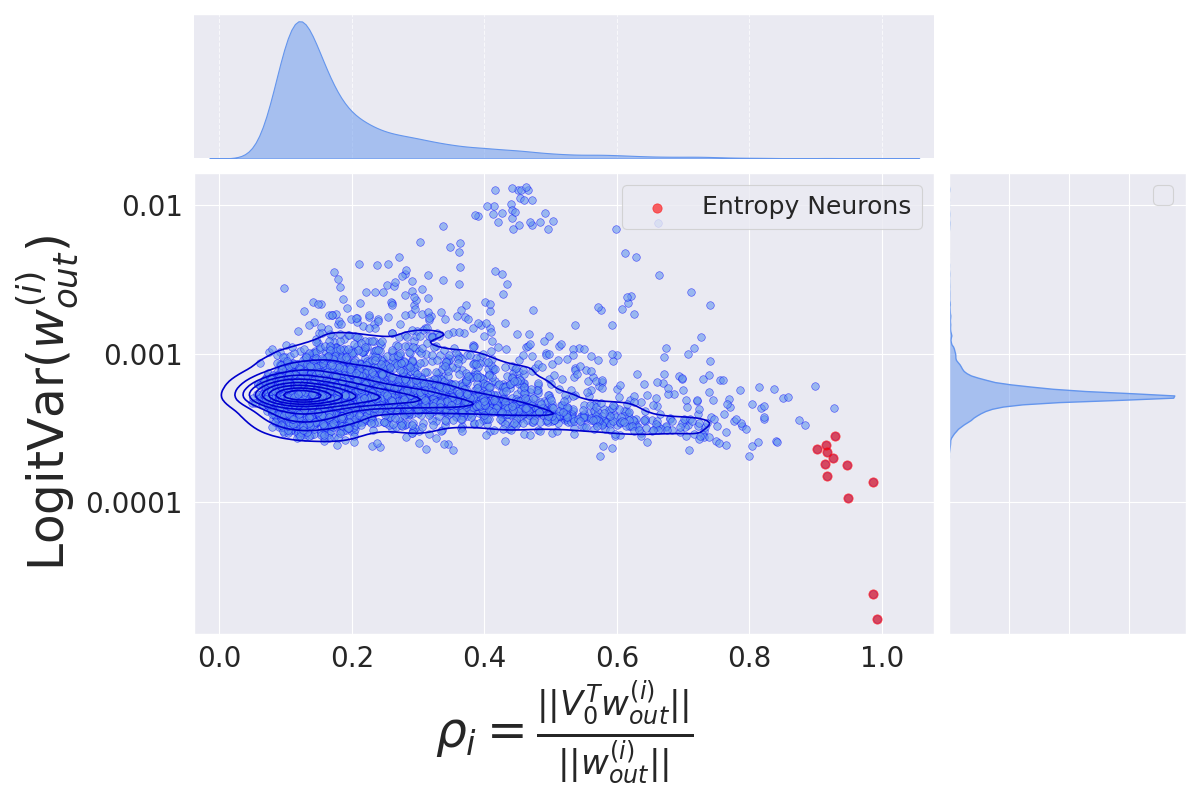}
    \caption{\textbf{Selected entropy neurons for Phi-1.5 (red).}}
    \label{fig:entropy_neurons_selection}
\end{figure}

\subsection{Entropy Neurons Selection}
\label{sec:entropy_neurons_selection}
We focus on the last Transformer layer because its entropy neurons have the most direct impact on the term logit distribution (through the projection with the unembedding matrix $\mathbf{W}_\mathrm{U}$). We use both LogitVar and $\rho$ (motivated by previous work on effective null space projections \citep{stolfo2024confidenceregulationneuronslanguage}) to select these neurons. 

Figure~\ref{fig:entropy_neurons_selection} \textcolor{black}{illustrates all the neurons with their corresponding} LogitVar and $\rho$ for Phi-1.5, with similar figures for \textcolor{black}{other models} in Figure~\ref{fig:entropy_neurons_selection_gpt_pythia_llama_mistral} in the Appendix. We select neurons with minimal logit variance impact (LogitVar) and \textcolor{black}{high projection} with $\mathbf{W}_\mathrm{U}$'s \textcolor{black}{effective null space} ($\rho$). For Phi-1.5, we select 12 entropy neurons, representing 1.5\textperthousand{} of the last layer's neurons, using a minimalist approach to pick the fewest neurons with strong characteristics. Table \ref{tab:model_dimensions} in the Appendix details hidden dimensions and selected entropy neuron proportions for all models.

\textcolor{black}{Although \citet{gurnee2024universal} and \citet{stolfo2024confidenceregulationneuronslanguage} observed high weight norm $||w_{\mathrm{out}}^{\mathrm{(i)}}||$ for entropy neurons (e.g., GPT-2, Figure \ref{fig:weigh_norm_distro_gpt2}) and used it to select entropy neurons}, we do not use high weight norm as a selection criterion. We observe that for some models, neurons with low LogitVar$(w_{\mathrm{out}}^{\mathrm{(i)}})$ and high $\rho_i$ can have relatively low $||w_{\mathrm{out}}^{\mathrm{(i)}}||$ compared to other neurons, as \textcolor{black}{illustrated} in Figure \ref{fig:weigh_norm_distro_llama3} for Llama-3-8B. Therefore, we consider LogitVar and $\rho$ as the crucial selection criteria.

\section{Mechanistic Study}
We present the metrics in \ref{subsection:measuring_neuron_impact}, and describe our results in \ref{subsection:results}.
\subsection{Metrics}
\label{subsection:measuring_neuron_impact}
We measure the impact of a set of neurons $\mathcal{N}$ on the context copying behavior by turning off these neurons, through causal interventions, and observing how the knowledge source (CK, PK or ND) changes (see Section \ref{subsection:framework} \textcolor{black}{and the schema in Figure \ref{fig:schema_methodo}}). In practice, \textcolor{black}{we turn off these neurons by replacing their activation values $n_i$} by an average value $\mu_{n_i}$ computed \textcolor{black}{over the knowledge probing dataset $E$}\footnote{\textcolor{black}{We also tested other ablation values and show their Global Transition Score in Table~\ref{tab:global_transition_ratio_by_ablation_value} in the Appendix.}}. More formally, \textcolor{black}{for each example \(x \in E\) (see Section~\ref{subsection:framework})}, \(K_{\mathcal{M}}(x)\) is the knowledge source used by the model \(\mathcal{M}\), and \(K_{\mathcal{M}\setminus \mathcal{N}}(x)\) is the knowledge source used by the ablated model \(\mathcal{M}^{\setminus \mathcal{N}}\) \textcolor{black}{given the input $x$.} 
Let $E_{K} = \{x \in E | K_{\mathcal{M}}(x) = K\}$ and $E_{\bar{K}} = E \setminus E_{K}$. 
We define the following metrics:


\begin{figure*}[t]
\centering
\begin{subfigure}{0.32\linewidth}
    \centering
    \includegraphics[width=\textwidth]{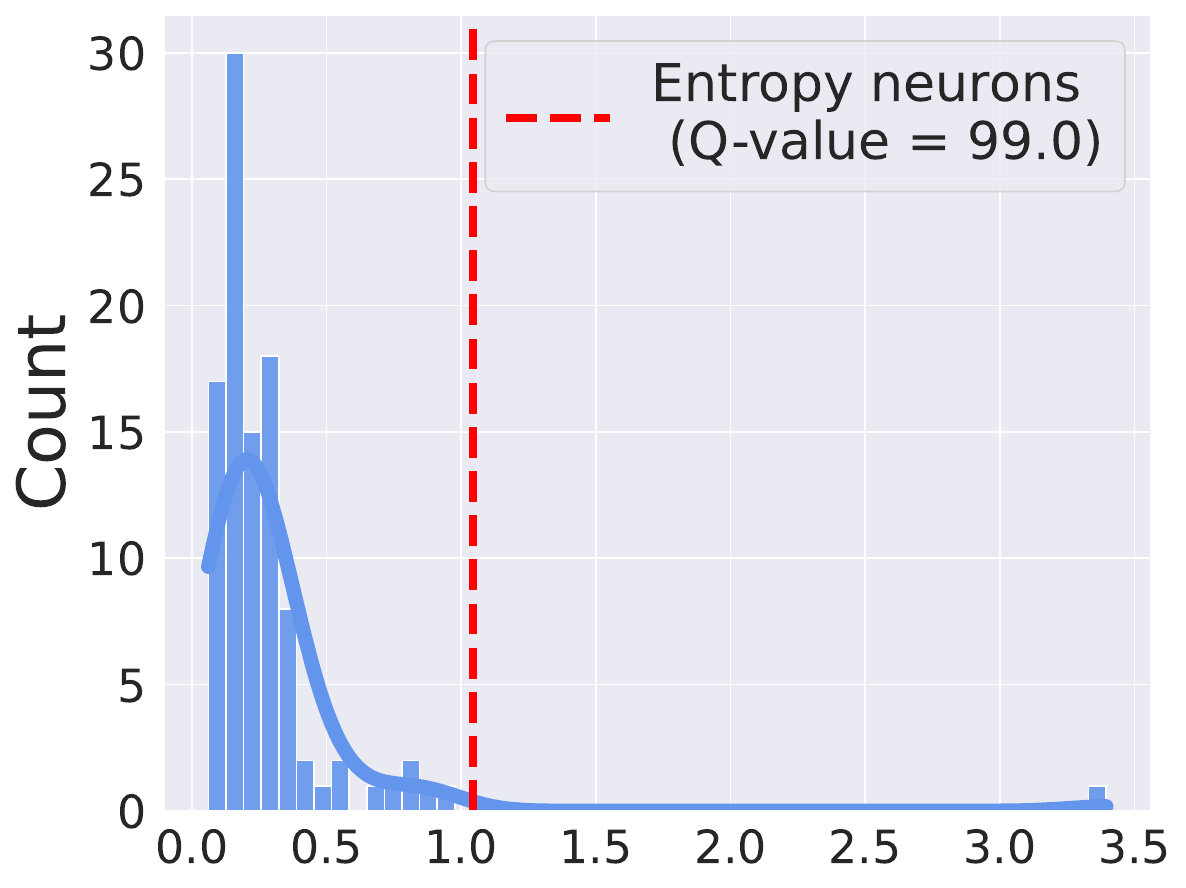}
    \caption{\textbf{Global Transition Score (\%)}}
    \label{fig:global_transition_ratio_phi}
\end{subfigure}
\hfill
\begin{subfigure}{0.32\linewidth}
    \centering
    \includegraphics[width=\textwidth]{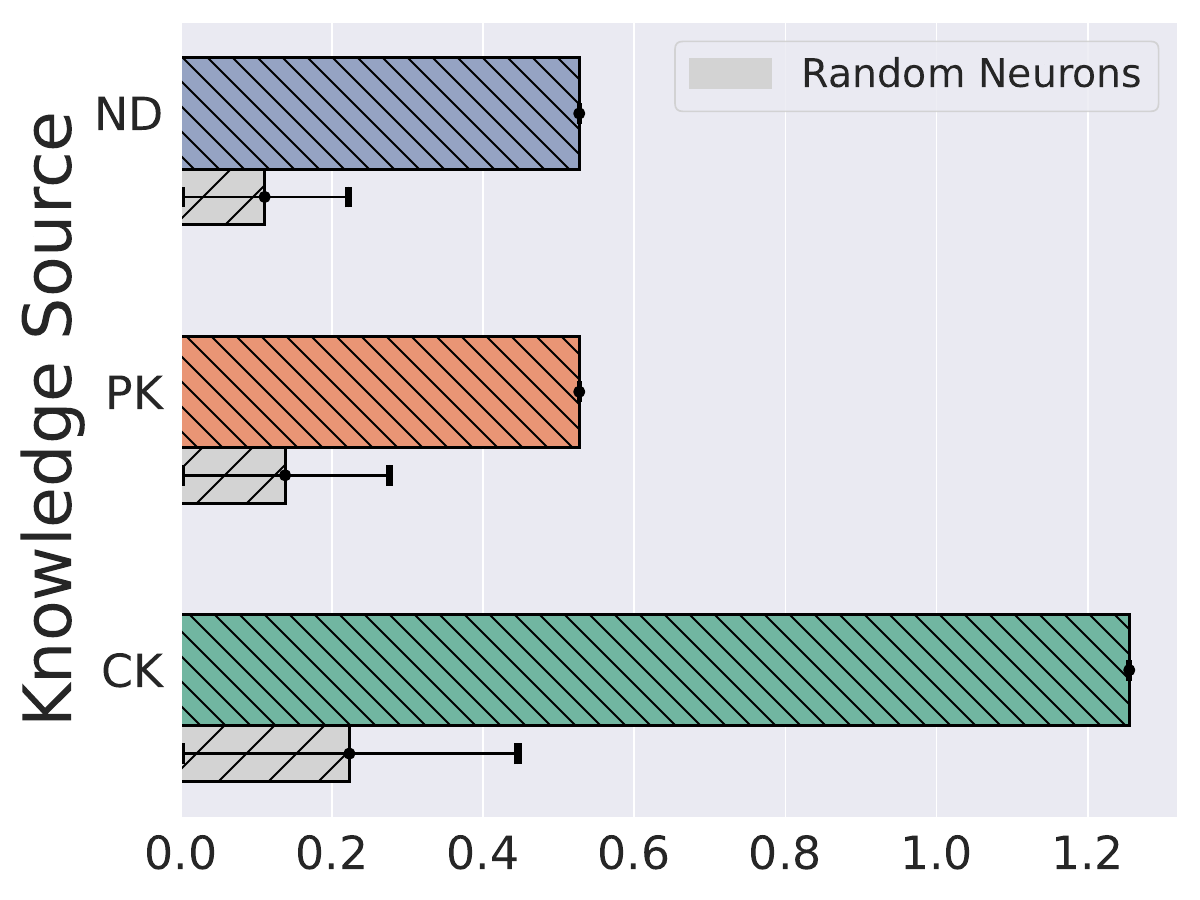}
    \caption{\textbf{Conversion Ratio (\%)}}
    \label{fig:conversion_ratio_phi}
\end{subfigure}
\hfill
\begin{subfigure}{0.32\linewidth}
    \centering
    \raisebox{1.55\height}{
    \resizebox{\linewidth}{!}{
    \begin{tabular}{l | ccc}
        \hline
        \textbf{} & \textbf{To CK} & \textbf{To ND} & \textbf{To PK} \\
        \hline
        \textbf{From CK} & \makecell{99.5\\ \scriptsize ($99.8\pm 0.1$) } & \makecell{0.2\\ \scriptsize ($0.1\pm 0.0$) } & \makecell{0.3\\ \scriptsize ($0.1\pm 0.1$) } \\
        \hline
        \textbf{From ND} & \ckconversion{\makecell{2.5\\ \scriptsize ($0.3\pm 0.1$) }} & \makecell{96.4\\ \scriptsize ($99.5\pm 0.1$) } & \makecell{1.1\\ \scriptsize ($0.2\pm 0.1$) } \\
        \hline
        \textbf{From PK} & \makecell{0.4\\ \scriptsize ($0.2\pm 0.1$) } & \makecell{0.1\\ \scriptsize ($0.1\pm 0.1$) } & \makecell{99.5\\ \scriptsize ($99.7\pm 0.1$) } \\
        \hline
    \end{tabular}
    }
    }
    \caption{\textbf{Transition Scores (\%)}}
    \label{table:mean_ablation_ablation_transition_table_phi}
\end{subfigure}
\caption{\textbf{Phi-1.5 ablation scores.} As a control, we provide the average Transition Score of 100 random ablations with its corresponding error bars ($\pm 3 \times$standard deviation). \textcolor{black}{We also provide the error bars for the entropy neurons in Figure~\ref{fig:conversion_ratio_phi} illustrated on top of the CK, PK, and ND bars.}}
\label{fig:phi_1_5_overview}
\end{figure*}

\paragraph{Global Transition Score (GTS):} proportion of examples for which the knowledge source changes as we remove the group of neurons $\mathcal{N}$

\small
\begin{equation}
\text{GTS} = \frac{1}{|E|} \sum_{x \in E} \mathbb{I}[K_{\mathcal{M}}(x) \neq K_{\mathcal{M}\setminus \mathcal{N}}(x)],
\end{equation}
\normalsize

\noindent where \(\mathbb{I}\) is the indicator function, equal to 1 if the condition is true and 0 otherwise, and \(|E|\) \textcolor{black}{is the cardinality of E}.
\textit{A high GTS indicates that ablating \(\mathcal{N}\) significantly alters the model's knowledge source selection}, underscoring the role of \(\mathcal{N}\) in knowledge source decision-making.

\paragraph{Conversion Ratio (CR):} proportion of examples \textcolor{black}{where the model switched} \emph{to} a given knowledge source $K \in \text{(PK, CK, ND)}$ when we remove $\mathcal{N}$

\small
\begin{equation}
\text{CR}(K) = \frac{1}{|E_{\bar{K}}|}\sum_{x \in E_{\bar{K}}} \mathbb{I}[ K_{\mathcal{M}\setminus \mathcal{N}}(x) = K]
\end{equation}
\normalsize


\noindent a high \(\text{CR}(K)\) suggests that ablating \(\mathcal{N}\) alters a large proportion  of examples towards $K$, indicating that \textit{\(\mathcal{N}\) is an inhibitor of the knowledge source $K$ in the original model \(\mathcal{M}\)}.
\vspace{-0.3cm}

\paragraph{Transition Score (TS):} proportion of examples that transition from knowledge source $K$ to knowledge source $K'$ as we remove $\mathcal{N}$ 

\vspace{-0.3cm}
\small
\begin{equation}
\text{TS}(K, K') = \frac{1}{|E_K|}\sum_{x \in E_K} \mathbb{I}[K_{\mathcal{M}\setminus \mathcal{N}}(x) = K']
,
\end{equation}
\normalsize

\noindent \textit{a high $\text{TS}(K, K')$} indicates that ablating $\mathcal{N}$ moves a large portion of examples with knowledge source $K$ to knowledge source $K'$, \textit{suggesting that the entropy neurons $\mathcal{N}$ tend to promote $K$ over $K'$.}
\subsection{Results}
\label{subsection:results}
\textcolor{black}{\textbf{Control Distribution:} to assess the significance of the results on entropy neurons $\mathcal{E}$, we build a control distribution by drawing 100 independent sets of neurons from the set of non-entropy neurons with the same cardinality as $\mathcal{E}$.}
\paragraph{Entropy neurons significantly influence the knowledge source of predictions.}
We investigated the impact of removing entropy neurons on knowledge source transitions (CK, PK, ND) across various models. Figure~\ref{fig:global_transition_ratio_phi} illustrates that ablating entropy neurons $\mathcal{E}$ results in a Global Transition Score (GTS) at the top 1\% of the control distribution for Phi-1.5. This suggests that entropy neurons play a significant role in the decision-making process regarding knowledge sources. This observation holds true for other models (see Figure~\ref{fig:gts_rest_of_models}).

\paragraph{Entropy neurons inhibit the induction mechanism.}
After demonstrating that removing entropy neurons triggers transitions between knowledge sources, we further analyzed the destination of these transitions using the Conversion Ratio (CR($K$)). Figure~\ref{fig:conversion_ratio_phi} for Phi-1.5, show a high CR for CK compared to the control distribution, indicating a significant shift from PK and ND to CK (highlighted in green) after ablating $\mathcal{E}$. This finding is corroborated by the Transition Scores presented in Table~\ref{table:mean_ablation_ablation_transition_table_phi} for Phi-1.5 (2.5\%) and in Table~\ref{table:mean_ablation_transition_table} (Appendix) for Llama-3-8B (6.2\%), GPT-2 (3.3\%), and Pythia-1.4B (2\%). We show in Table \ref{tab:pk_to_ck_switch_examples} examples where Phi-1.5 switched from using PK to CK.


\section{Conclusion}
In this paper, we demonstrated that entropy neurons play a significant role in modulating the balance between PK and CK. Ablation studies revealed that perturbing these neurons leads to significant shifts in the knowledge source used by the model. Specifically, the GTS for entropy neurons is at the top 1\% of the control distribution, this finding is consistent for different models up to 8B parameters. 
\textcolor{black}{More broadly, identifying entropy neurons as inhibitors of context copying contributes to a clearer picture of how LLMs manage conflicting sources of knowledge. This lays the groundwork for future work on characterizing and interpreting the internal dynamics of LLMs, and more specifically helping to explain when and why models rely on PK vs CK.}

\section{Limitations}
\label{sec:limitations}
While our experiments demonstrate that entropy neurons significantly inhibit context copying behavior in LLMs, our study is limited by an incomplete understanding of the broader copying regulation mechanism. Specifically, we focused solely on entropy neurons in the FFN of the final transformer layer, which may neglect the contributions of other neuron types and architectural components.

Additionally, \textcolor{black}{although we observed relatively high Global Transition Scores in most of the models we studied, their Q-values varies. For instance, in Phi-1.5, Llama-3-8B, and GPT-2 the Q-value is around 99 which is less for Mistral-7B-v0.1 and Pythia-1.4B with 91 and 92.5 respectively. Model architecture and training could explain this variation.}

\textcolor{black}{Lastly, our study focuses on how entropy neurons contribute to modulating the balance between parametric and contextual knowledge in a situation of induction and does not explore why this specific set of neurons act this way.}

Future research should therefore expand the investigation to include a wider array of neural components and alternative perturbation methods to more comprehensively elucidate the underlying processes governing copying regulation. \textcolor{black}{It should also explore the reasons why entropy neurons specifically contribute to modulating the balance between CK and PK in situations of induction.} It could also be interesting to explore the role of these components on other general linguistic tasks \cite{tighidet-ballier-2022-fine, kaddour2023challengesapplicationslargelanguage}.

\section{Ethical Considerations}
Our study probes the internal mechanisms of large language models (LLMs) by manipulating a small subset of neurons—entropy neurons—that modulate the balance between parametric and contextual knowledge. All experimental data and prompts are derived from publicly available sources minimizing any direct privacy or security concerns.

However, we acknowledge that our findings have some implications. The probing and ablation techniques we describe could be repurposed to intentionally bias or subvert LLM behavior. Specifically, the structured prompts we employ to induce context copying may serve as templates for adversarial attacks, allowing malicious actors to manipulate model outputs in subtle but impactful ways. Similarly, our demonstration that targeted neuron ablation alters a model’s decision-making process raises the risk that LLMs could be engineered—intentionally or inadvertently—to prioritize deceptive or harmful outputs.

Given these risks, we stress the importance of applying this work within responsible and well-governed research contexts. We urge future researchers to incorporate safeguards against misuse, including robust evaluation pipelines and transparency in experimental intent. To foster reproducibility and critical engagement, we have released our codebase under an open license while documenting the limitations of our approach. 

\bibliography{custom}

\appendix

\section{Hardware and Software}

Experiments were performed using NVIDIA H100 and A100 GPUs, each equiped with 80 GB of VRAM. The process of generating the outputs with and without ablations took around 250 GPU hours. Our codebase was built using PyTorch \cite{paszke2019pytorch}, the HuggingFace Transformers library \cite{wolf-etal-2020-transformers} the TransformerLens library \cite{nanda2022transformerlens}, and the knowledge probing framework \cite{tighidet2024probinglanguagemodelsknowledge}.

\section{License}

Llama3-8B weights are released under the license available at \url{https://llama.meta.com/llama3/license/}. Mistral-7B and Pythia-1.4B weights are released under an Apache 2.0 license. Phi-1.5 and GPT-2 weights are released under a MIT license.

\section{Weight Pre-processing}
\label{weight_preprocessing}
To eliminate irrelevant components and other parameterization degrees of freedom, we utilize a set of standard weights pre-processing techniques following~\citet{nanda2022transformerlens} and \citet{stolfo2024confidenceregulationneuronslanguage}.

\textbf{Incorporating Layer Norm.} Most layer norm implementations include trainable parameters $\gamma \in \mathbb{R}^n$ and $\beta \in \mathbb{R}^n$. To account for these, we "fold" the layer norm parameters into $\mathbf{W}_{\mathrm{in}}$ by treating the layer norm parameters as equivalent to a linear layer and then combining the adjacent linear layers. We create effective weights as follows:
\begin{align}
    \mathbf{W}_{\mathrm{eff}} = \mathbf{W}_{\mathrm{in}} \cdot \text{diag}(\gamma), \quad \beta_{\mathrm{eff}} = \beta_{\mathrm{in}} + \mathbf{W}_{\mathrm{in}} \cdot \beta
\end{align}
\vspace{-0.05cm}
Finally, we center the reading weights because the preceding layer norm projects out the all-ones vector. Thus, we center the weights $\mathbf{W}_{\mathrm{eff}}$ as follows:
\begin{align}
    \mathbf{W}_{\mathrm{eff}}^{'}(i, :) = \mathbf{W}_{\mathrm{eff}}(i, :) - \mathbf{\bar{W}}_{\mathrm{eff}}(i, :).
\end{align}

\textbf{Centering Writing Weights.} Every time the model interacts with the residual stream, it applies a LayerNorm first. Therefore, the components of $\mathbf{W}_{\mathrm{out}}$ and $\beta_{\mathrm{out}}$ that lie along the all-ones direction of the residual stream have no effect on the model's calculations. Consequently, we mean-center $\mathbf{W}_{\mathrm{out}}$ and $\beta_{\mathrm{out}}$:
\begin{align}
    \mathbf{W}_{\mathrm{out}}^{'} = \mathbf{W}_{\mathrm{out}}(:, i) - \mathbf{\bar{W}}_{\mathrm{out}}(:, i).
\end{align}

\textbf{Centering Unembedding.} Since softmax is translation invariant, we also center $\mathbf{W}_\mathrm{U}$:
\begin{align}
    \mathbf{W}_\mathrm{U}^{'}(:, i) = \mathbf{W}_\mathrm{U}(:, i) - \mathbf{\bar{W}}_\mathrm{U}(:, i)
\end{align}

\section{Data Characteristics}
\label{sec:data_characteristics}
\begin{figure}[H]
	\begin{centering}\includegraphics[width=0.8\columnwidth]{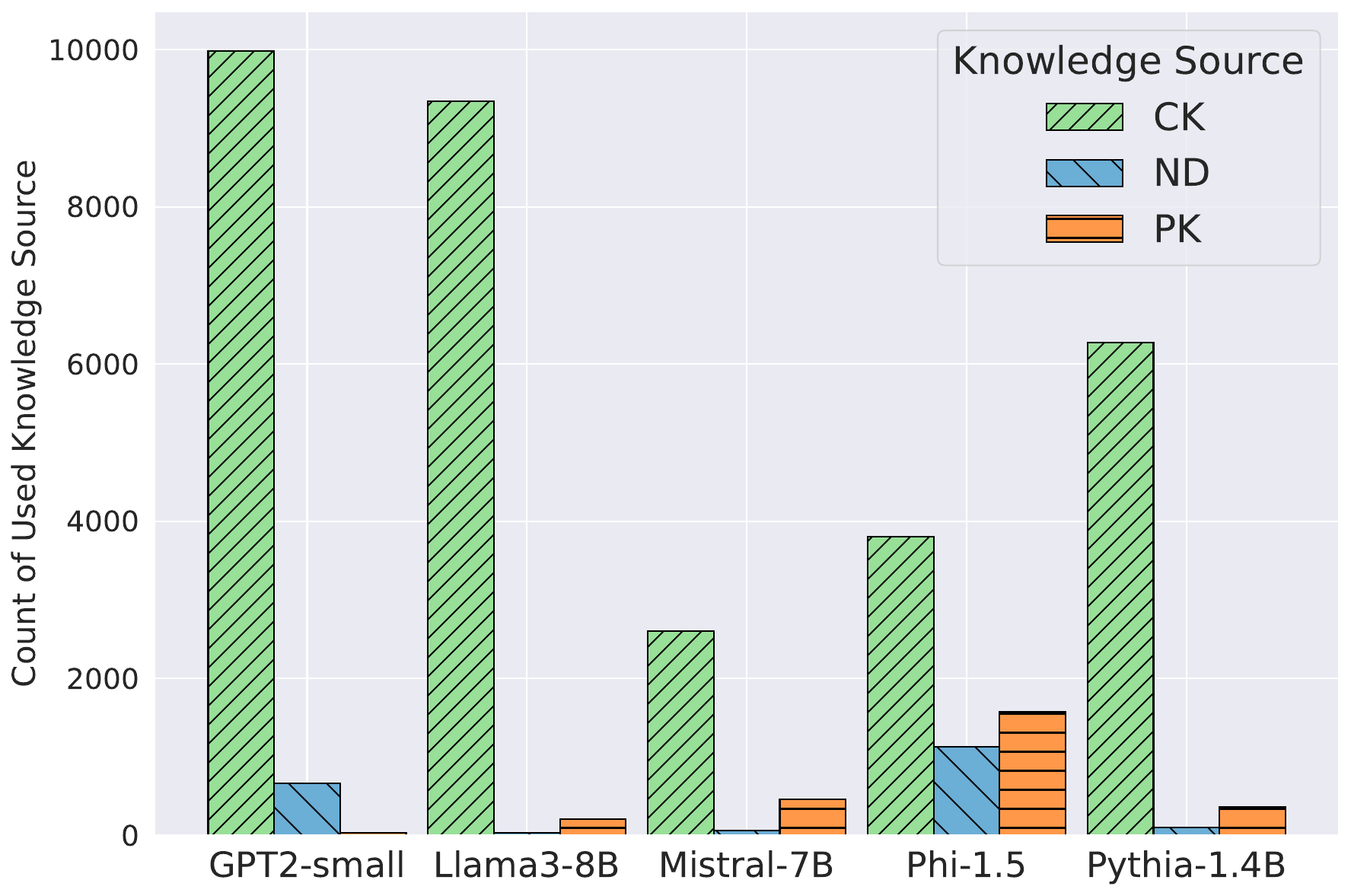}\caption{\textbf{Count of used knowledge sources by each model before ablation.}}
	\label{fig:knowledge_source_count}
	\end{centering}
\end{figure}

We provide in Figure \ref{fig:knowledge_source_count} the count of used knowledge sources by model before ablating entropy neurons. We also provide in Table \ref{table:prompting_sample} a sample of examples from the knowledge probing dataset.
\section[W\_U's \textcolor{black}{Effective Null Space}]{$\mathbf{W}_\mathrm{U}$'s \textcolor{black}{Effective Null Space}}
\label{sec:null_space_identification}
\begin{figure}[H]
    \centering
    \begin{subfigure}[b]{0.49\columnwidth}
        \centering
        \includegraphics[width=\columnwidth]{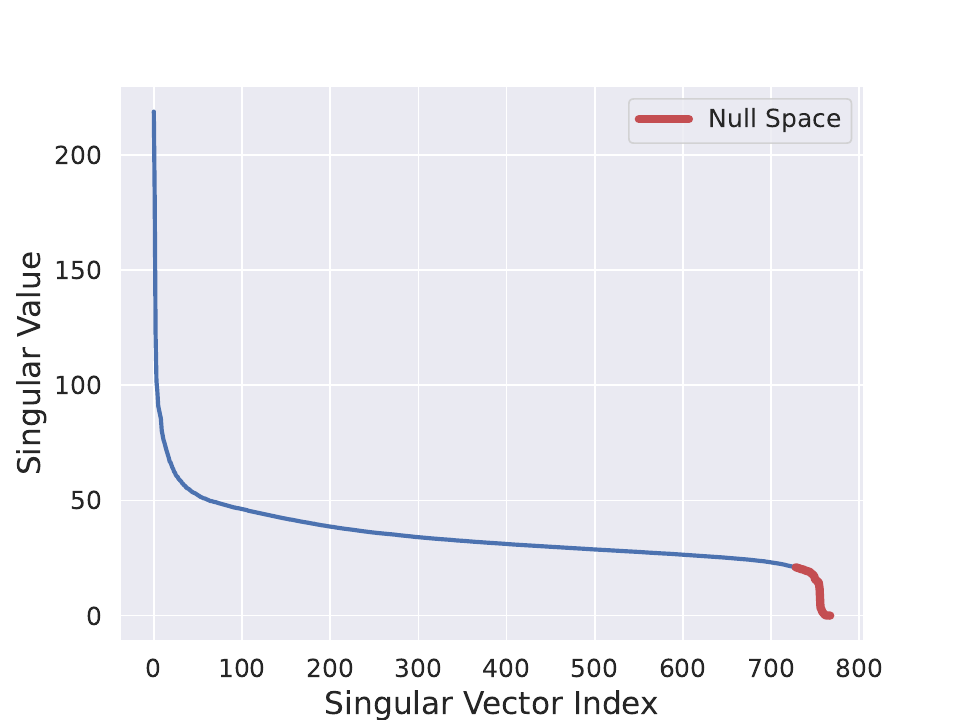}
        \caption{GPT2}
    \end{subfigure}
    \begin{subfigure}[b]{0.49\columnwidth}
        \centering
        \includegraphics[width=\columnwidth]{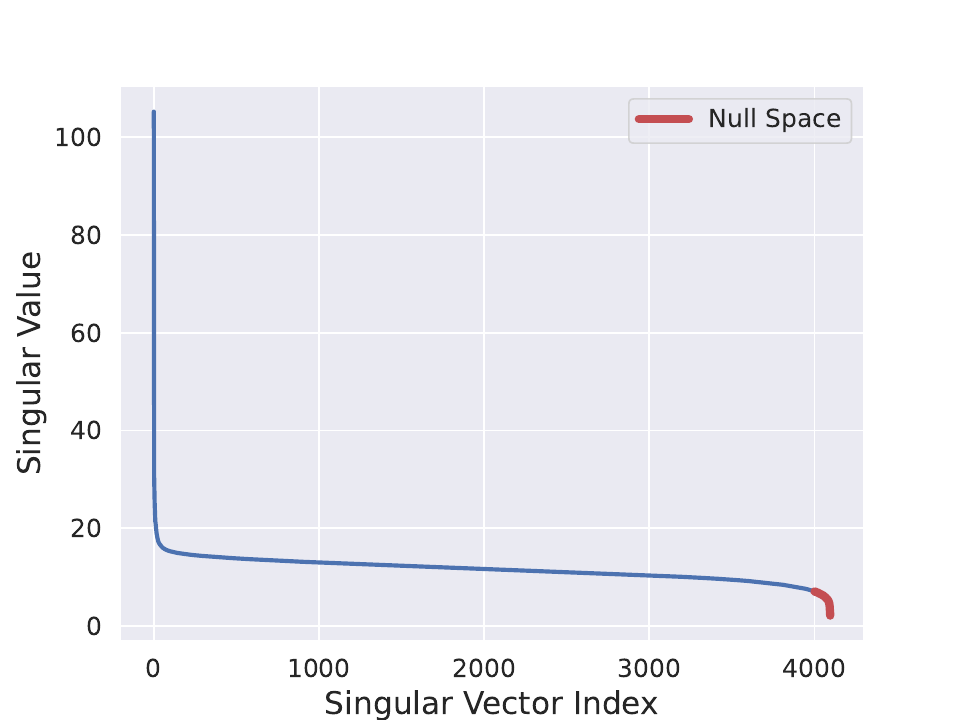}
        \caption{Llama-3-8B}
    \end{subfigure}
    \begin{subfigure}[b]{0.49\columnwidth}
        \centering
        \includegraphics[width=\columnwidth]{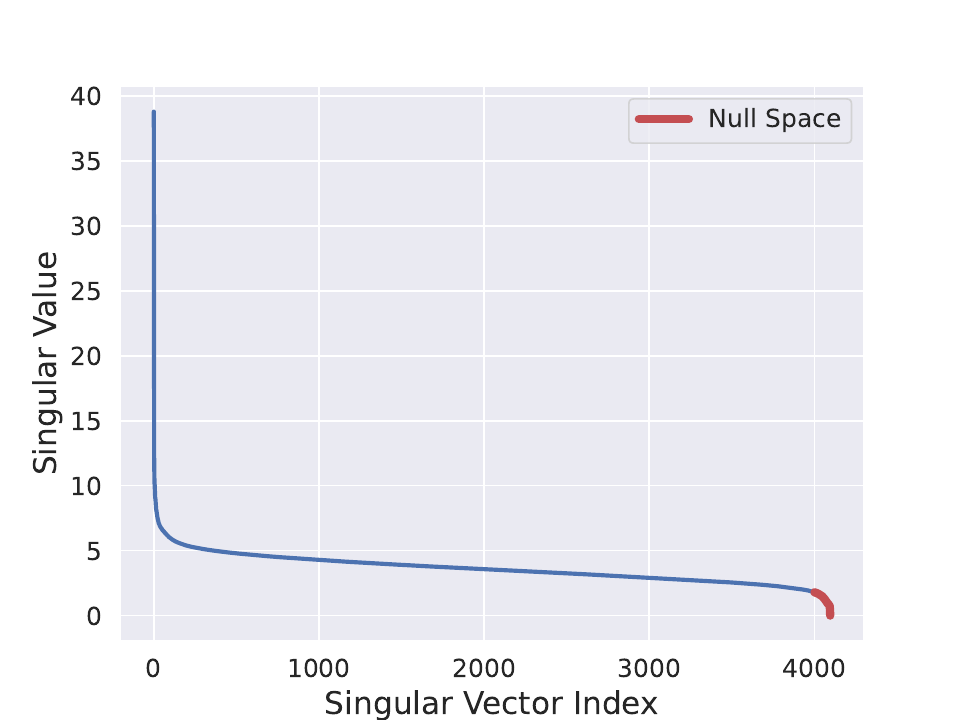}
        \caption{Mistral-7B-v0.1}
        \label{fig:sing_values_null_space_mistral}
    \end{subfigure}
    \begin{subfigure}[b]{0.49\columnwidth}
        \centering
        \includegraphics[width=\columnwidth]{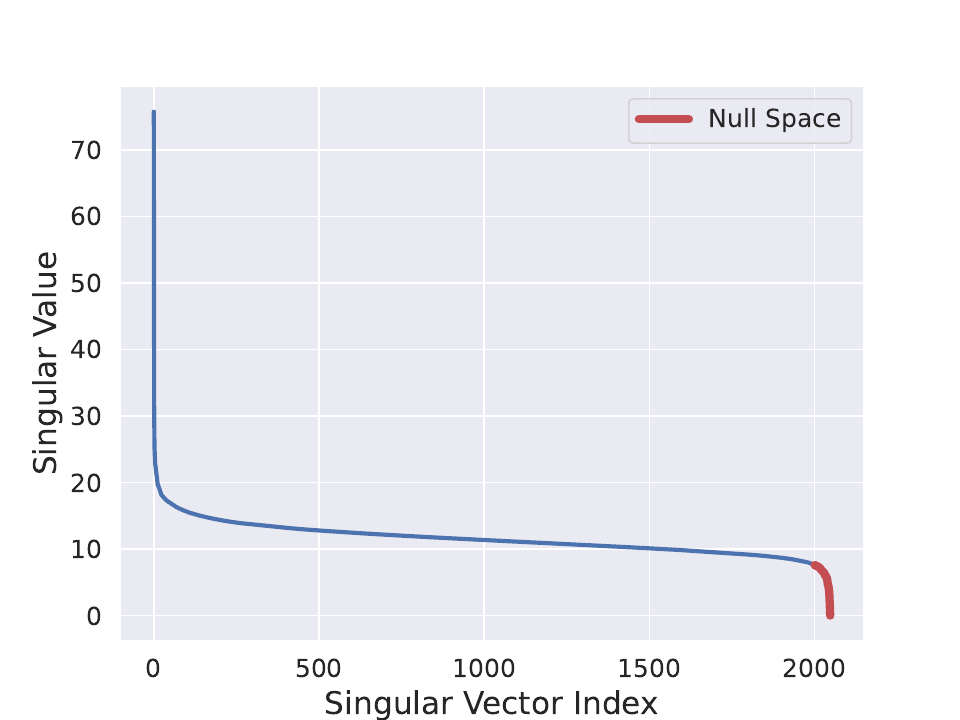}
        \caption{Pythia-1.4B}
    \end{subfigure}
    \begin{subfigure}[b]{0.49\columnwidth}
        \centering
        \includegraphics[width=\columnwidth]{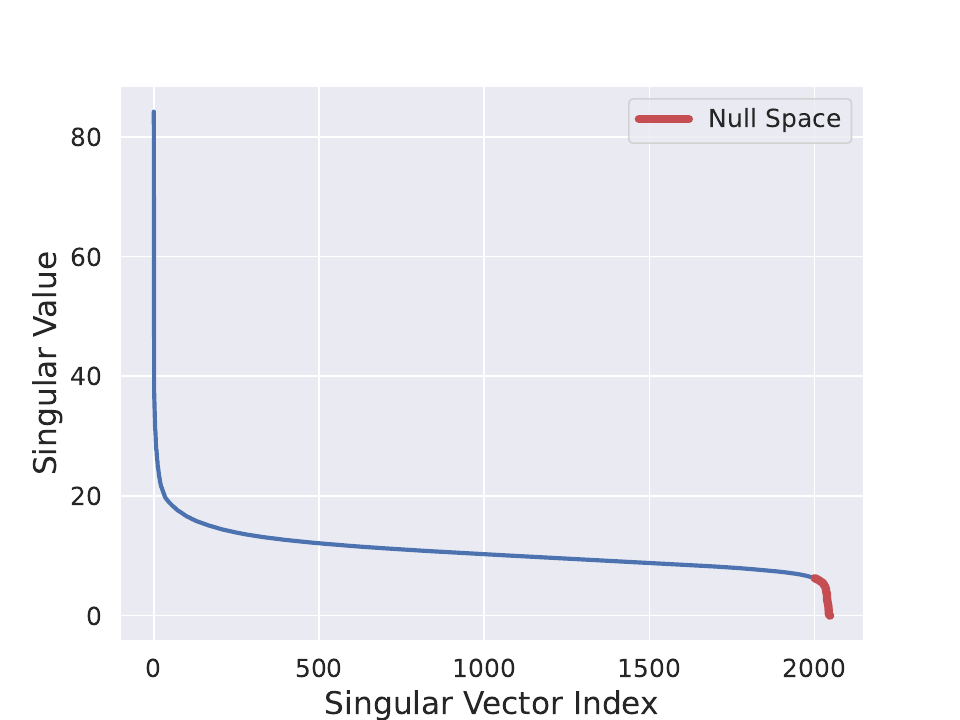}
        \caption{Phi-1.5}
    \end{subfigure}
    \caption{Unembedding matrix $\mathbf{W_U}$ singular values, illustrating the \textcolor{black}{effective null space} of $\mathbf{W_U}$ in red.}
    \label{fig:sing_values_null_space_all_models}
\end{figure}

To identify the \textcolor{black}{effective null space} $\mathbf{V_0}$ of $\mathbf{W_U}$, we start by applying a singular value decomposition (\textbf{SVD}) on $\mathbf{W_U}$:
\begin{align}
    \textbf{SVD}(\mathbf{W_U}) = \mathbf{U} \Sigma \mathbf{V^T},
\end{align}
we then consider the right singular vectors with the lowest singular values, noted $\mathbf{V_0}$, starting from a sharp drop as shown in Figure \ref{fig:sing_values_null_space_all_models}. We also detail the \textcolor{black}{effective null space} dimension size for all the studied models in Table \ref{tab:model_dimensions}.
\label{sec:null_space}

\section{Activations}

\begin{table}[h]
\small
\centering
\begin{tabular}{|l|l|l|}
\hline
\textbf{Model}           & \textbf{Activation Function} & \textbf{Domain} \\ \hline
Llama-3-8B                & SwiGLU: Swish $\times$ GLU   & $\mathbb{R}$    \\ \hline
Mistral-7B-V0.1          & SwiGLU: Swish $\times$ GLU   & $\mathbb{R}$    \\ \hline
Phi-1.5                  & GELU & $\mathbb{R}$    \\ \hline
Pythia-1.4B              & GELU & $\mathbb{R}$   \\ \hline
GPT-2-Small              & GELU & $\mathbb{R}$    \\ \hline
\end{tabular}
\caption{FFN hidden layer activation functions for all the studied models}
\label{table:activation_functions}
\end{table}

\begin{table*}[t]
\scriptsize
\centering
\resizebox{\textwidth}{!}{
\begin{tabular}{>{\raggedright\arraybackslash}p{9cm}ccc} 
    \toprule
    \textbf{Input Prompt} & \textbf{Knowledge Source} & \textbf{PK Attribute} & \textbf{Language Model} \\
    \midrule
    \rowcolor{lightblue} \textit{Harney County has its capital city in \underline{Taiwan}. Harney County has its capital city in \textbf{Burns.}} & ND & Oregon & Llama3-8B \\ 
    \midrule
    \rowcolor{lightblue} \textit{Lisa Appignanesi has citizenship of \underline{Finland}. Lisa Appignanesi has citizenship of \textbf{France.}} & ND & the UK & Llama3-8B \\ 
    \midrule
    \rowcolor{lightblue} \textit{Craiova is located in the continent of \underline{India}. Craiova is located in the continent of \textbf{Romania.}} & ND & \textit{Europe} & Pythia-1.4B \\ 
    \midrule
    \rowcolor{lightorange} \textit{The Kingdom of Hungary had its capital as \underline{Connecticut}. The Kingdom of Hungary had its capital as \textbf{Connecticut.}} & CK & \textit{Budapest} & Mistral-7B \\ 
    \midrule
    \rowcolor{lightorange} \textit{The Wii U system software is a product that was manufactured by \underline{Square}. The Wii U system software is a product that was manufactured by \textbf{Square.}} & CK & \textit{Nintendo} & Llama3-8B \\ 
    \midrule
    \rowcolor{lightorange} \textit{The Centers for Disease Control and Prevention is headquartered in \underline{Lyon}. The Centers for Disease Control and Prevention is headquartered in \textbf{Lyon.}} & CK & \textit{Atlanta} & Llama3-8B \\ 
    \midrule
    \rowcolor{lightgreen} \textit{Harare is the capital city of \underline{Florida}. Harare is the capital city of \textbf{Zimbabwe.}} & PK & \textit{Zimbabwe} & Pythia-1.4B \\ 
    \midrule
    \rowcolor{lightgreen} \textit{Goodreads is owned by \underline{Microsoft}. Goodreads is owned by \textbf{Amazon.}} & PK & \textit{Amazon} & Phi-1.5 \\ 
    \midrule
    \rowcolor{lightgreen} \textit{OneDrive is owned by \underline{Toyota}. OneDrive is owned by \textbf{Microsoft.}} & PK & \textit{Microsoft} & Mistral-7B \\ 
    \bottomrule
\end{tabular}
}
\caption{Examples of final probing prompts, including their knowledge source, the LLM, and the corresponding parametric knowledge (PK) object. Bold text indicates the generated attribute, while underlined text represents the counter-knowledge attribute.}
\label{table:prompting_sample}
\end{table*}
\begin{table*}[t]
    \centering
    \small
    \begin{tabular}{lcccccc}
        \toprule
        \textbf{Model} & \textbf{$d_{model}$} & \textbf{$d_{\text{ffn}}$} & \textbf{$d_{\text{\textcolor{black}{effective null space}}}$} & \textbf{Card($V$)} & \textbf{$\frac{d_{\text{\textcolor{black}{effective null space}}}}{d_{model}}$ (\%)} & \makecell{Entropy Neurons\\ (\textperthousand)} \\
        \midrule
        GPT-2 & 768 & 3072 & 40 & 50257 & 5.20 & 2 \\
        Llama-3-8B & 4096 & 14336 & 96 & 128256 & 2.34 & 0.7 \\
        Mistral-7B-v0.1 & 4096 & 14336 & 96 & 32000 & 2.34 & 1 \\
        Pythia-1.4B & 2048 & 8192 & 48 & 50304 & 2.34 & 1.1 \\
        Phi-1.5 & 2048 & 8192 & 48 & 51200 & 2.34 & 1.5 \\
        \bottomrule
    \end{tabular}
    \caption{Models hidden dimensions compared to the proportion of selected entropy neurons.}
    \label{tab:model_dimensions}
\end{table*}
%
\begin{figure*}[t]
    \centering
    \begin{subfigure}[b]{0.49\textwidth}
        \centering
        \includegraphics[width=\textwidth]{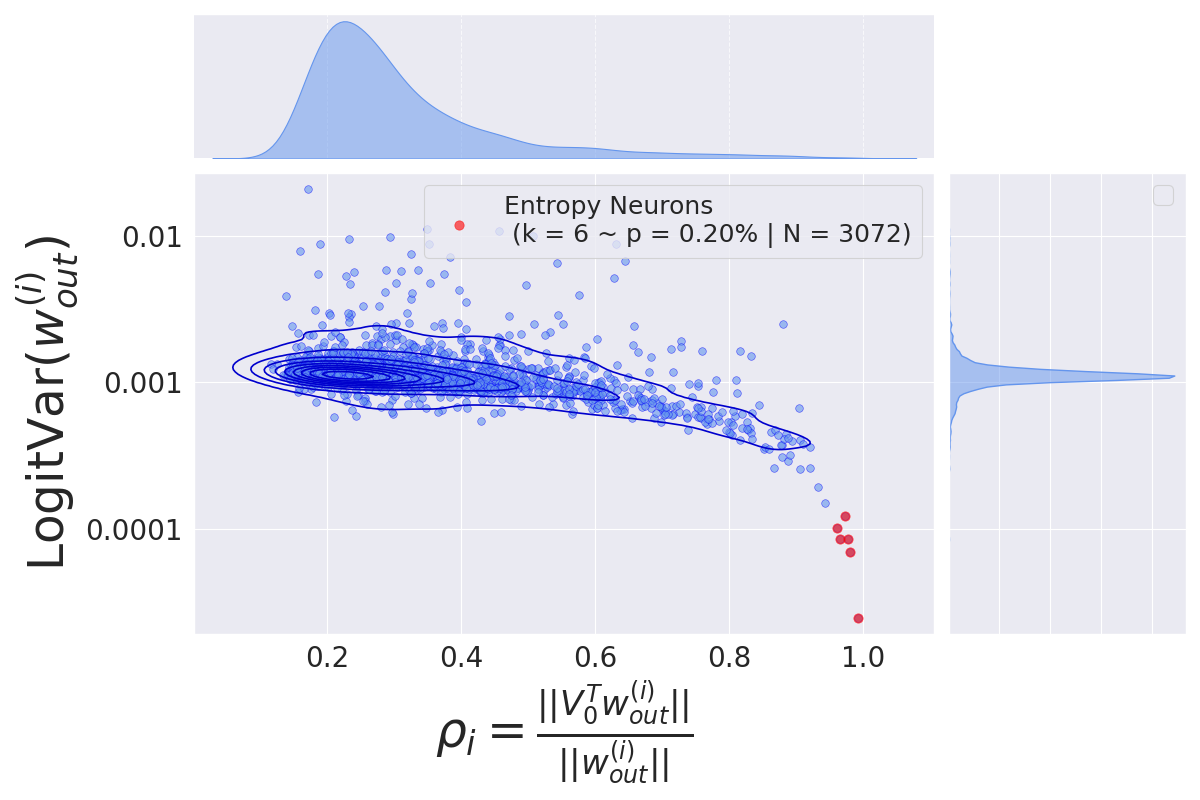}
        \caption{GPT2}
    \end{subfigure}
    \begin{subfigure}[b]{0.49\textwidth}
        \centering
        \includegraphics[width=\textwidth]{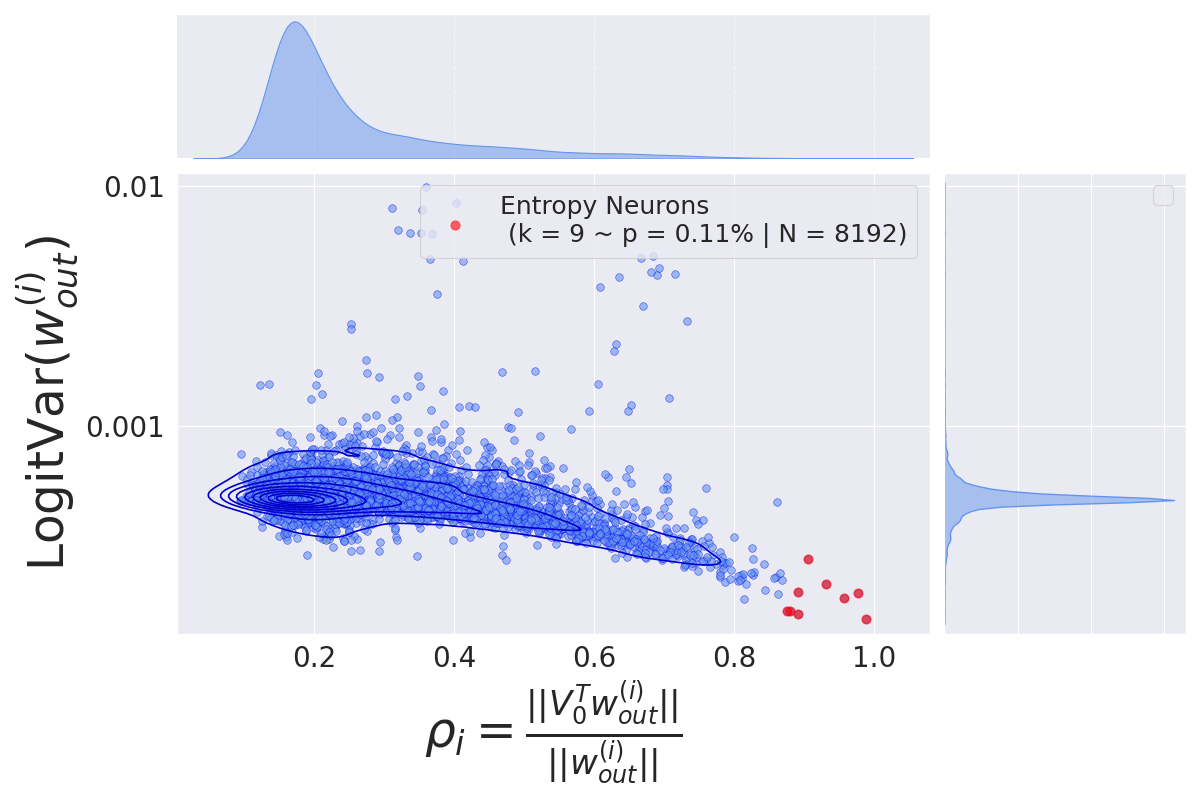}
        \caption{Pythia-1.4B}
    \end{subfigure}
    \\
    \begin{subfigure}[b]{0.49\textwidth}
        \centering
        \includegraphics[width=\textwidth]{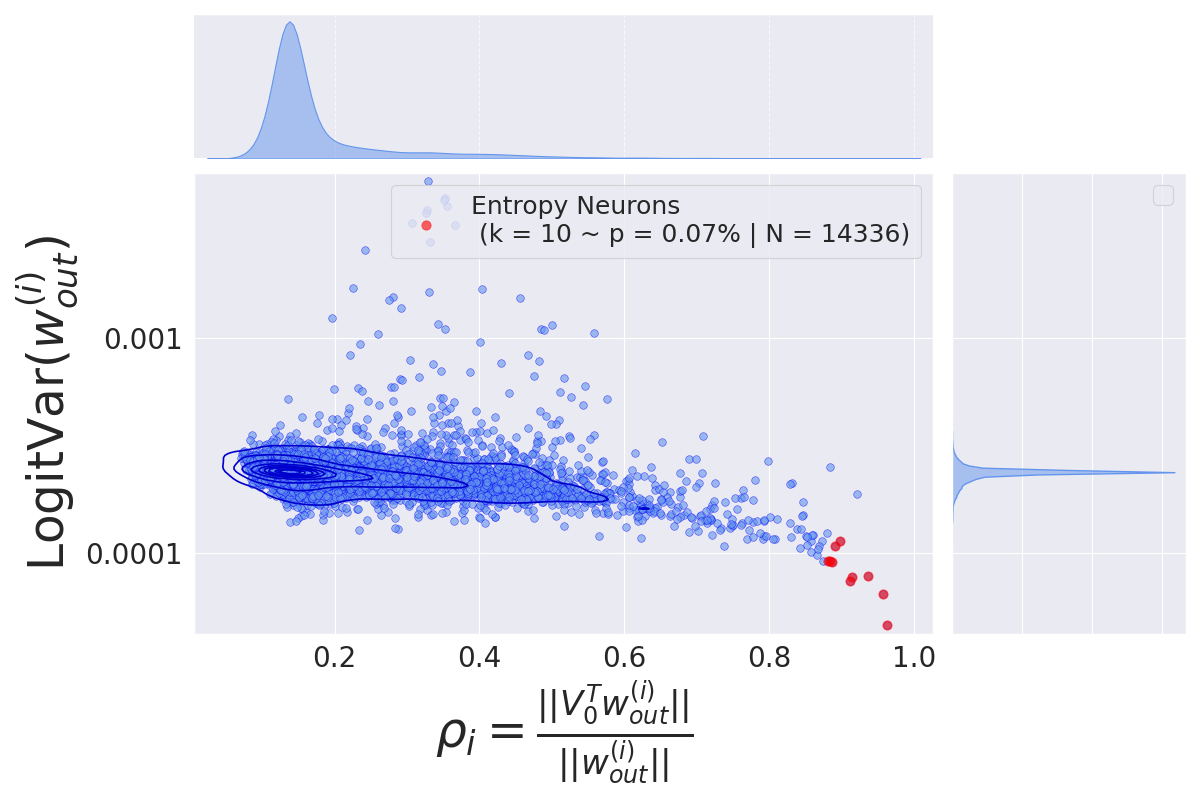}
        \caption{Llama-3-8B}
    \end{subfigure}
    \begin{subfigure}[b]{0.49\textwidth}
        \centering
        \includegraphics[width=\textwidth]{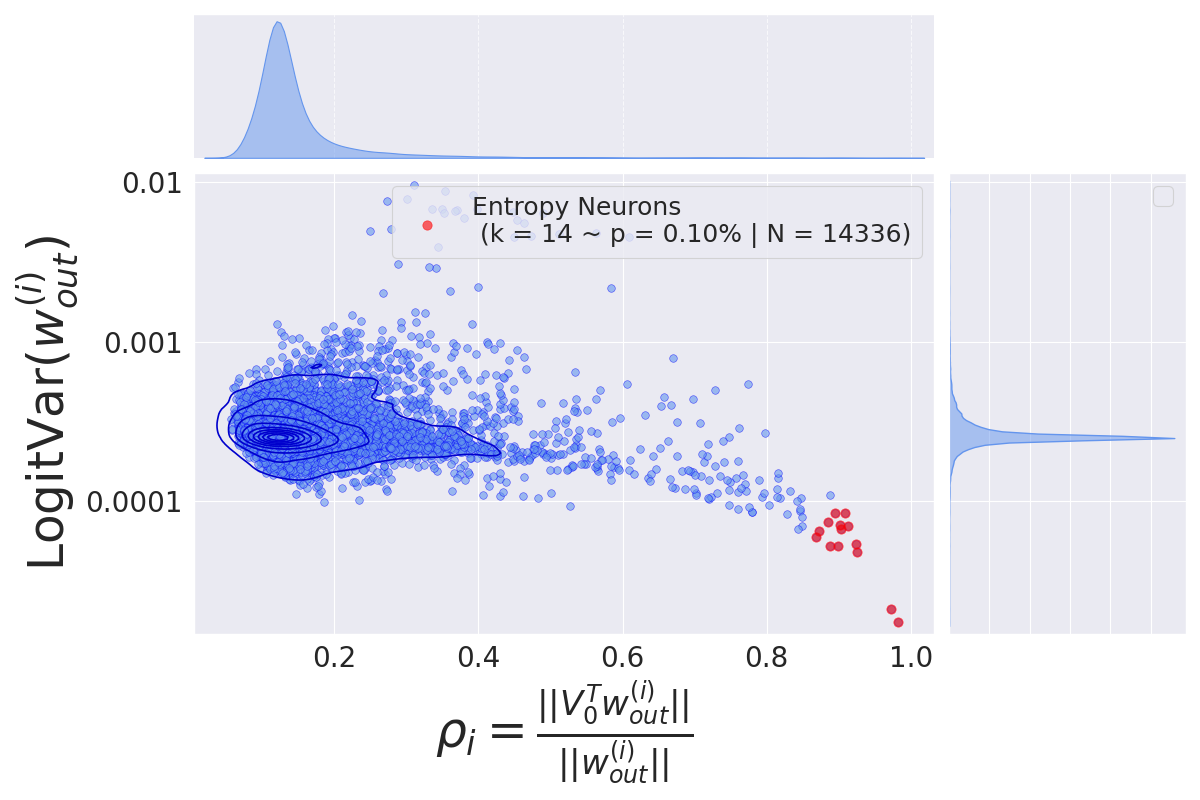}
        \caption{Mistral-7B}
    \end{subfigure}
    \caption{\textbf{Selected entropy neurons (red).} We select entropy neurons following the LogitVar and $\rho$ criteria. In each Figure, $k$ is the number of selected entropy neurons, $p$ is the proportions of entropy neurons, and $N$ is the total number of neurons.}
    \label{fig:entropy_neurons_selection_gpt_pythia_llama_mistral}
\end{figure*}%

\begin{table*}[t]
\footnotesize
\resizebox{\textwidth}{!}{
\begin{tabular}{l | ccc | ccc | ccc}
    \hline
    \rowcolor[gray]{0.95}
    \textbf{Model Name} & \multicolumn{3}{c|}{\textbf{From CK}} & \multicolumn{3}{c|}{\textbf{From ND}} & \multicolumn{3}{c}{\textbf{From PK}} \\
    \rowcolor[gray]{0.95} & \textbf{To CK} & \textbf{To ND} & \textbf{To PK} & \textbf{To CK} & \textbf{To ND} & \textbf{To PK} & \textbf{To CK} & \textbf{To ND} & \textbf{To PK} \\
    \hline
    \makecell{GPT-2} & \makecell{100.0\\ \scriptsize ($100.0\pm 0.0$) }& \makecell{0.0\\ \scriptsize ($0.0\pm 0.0$) }& \makecell{0.0\\ \scriptsize ($0.0\pm 0.0$) }& \makecell{3.3\\ \scriptsize ($0.4\pm 0.1$) }& \makecell{96.4\\ \scriptsize ($99.6\pm 0.1$) }& \makecell{0.3\\ \scriptsize ($0.0\pm 0.0$) }& \makecell{0.0\\ \scriptsize ($1.2\pm 0.6$) }& \makecell{6.2\\ \scriptsize ($2.6\pm 0.8$) }& \makecell{93.8\\ \scriptsize ($96.3\pm 1.0$) }\\ \hline
    \renewcommand{\arraystretch}{2.5}
    \makecell{Mistral-7B} & \makecell{99.8\\ \scriptsize ($99.9\pm 0.0$) }& \makecell{0.0\\ \scriptsize ($0.0\pm 0.0$) }& \makecell{0.2\\ \scriptsize ($0.1\pm 0.0$) }& \makecell{0.0\\ \scriptsize ($0.3\pm 0.3$) }& \makecell{98.6\\ \scriptsize ($99.3\pm 0.5$) }& \makecell{1.4\\ \scriptsize ($0.4\pm 0.3$) }& \makecell{2.2\\ \scriptsize ($0.6\pm 0.2$) }& \makecell{0.2\\ \scriptsize ($0.0\pm 0.0$) }& \makecell{97.6\\ \scriptsize ($99.4\pm 0.2$) }\\ \hline
    \renewcommand{\arraystretch}{2.5}
    \makecell{Llama3-8B} & \makecell{99.6\\ \scriptsize ($100.0\pm 0.0$) }& \makecell{0.1\\ \scriptsize ($0.0\pm 0.0$) }& \makecell{0.4\\ \scriptsize ($0.0\pm 0.0$) }& \makecell{6.2\\ \scriptsize ($0.2\pm 0.3$) }& \makecell{90.6\\ \scriptsize ($99.7\pm 0.4$) }& \makecell{3.1\\ \scriptsize ($0.1\pm 0.2$) }& \makecell{0.5\\ \scriptsize ($0.9\pm 0.3$) }& \makecell{0.5\\ \scriptsize ($0.0\pm 0.0$) }& \makecell{99.1\\ \scriptsize ($99.1\pm 0.3$) }\\ \hline
    \renewcommand{\arraystretch}{2.5}
    \makecell{Pythia-1.4B} & \makecell{99.9\\ \scriptsize ($100.0\pm 0.0$) }& \makecell{0.0\\ \scriptsize ($0.0\pm 0.0$) }& \makecell{0.1\\ \scriptsize ($0.0\pm 0.0$) }& \makecell{2.0\\ \scriptsize ($0.7\pm 0.2$) }& \makecell{98.0\\ \scriptsize ($99.3\pm 0.3$) }& \makecell{0.0\\ \scriptsize ($0.0\pm 0.1$) }& \makecell{0.0\\ \scriptsize ($0.3\pm 0.1$) }& \makecell{0.0\\ \scriptsize ($0.0\pm 0.0$) }& \makecell{100.0\\ \scriptsize ($99.7\pm 0.1$) }\\ \hline
\end{tabular}
}
\caption{Transition Scores (\%) From \text{source} To \text{target} knowledge source after mean ablating entropy neurons across models. As a control, we provide the average Transition Score of 100 random ablations with its corresponding error bars ($\pm 3\sigma$).}
\label{table:mean_ablation_transition_table}
\end{table*}
\begin{table*}[t]
    \centering
    \resizebox{0.8\textwidth}{!}{%
    \begin{tabular}{l l S[table-format=2.1] S[table-format=3.1]}
    \toprule
    \rowcolor[gray]{0.95}\textbf{Ablation Value} & \textbf{Model} & {\textbf{EN Transition Score (\%)}} & {\textbf{Q-val}} \\
    \midrule
    \multirow{5}{*}{$\mu_{n_i}$}
        & GPT-2 & 0.3 & 98.0 \\
        & Pythia-1.4B & 0.1 & 92.5 \\
        & Mistral-7B-v0.1 & 0.5 & 91.0 \\
        & Phi-1.5 & 1.0 & 99.0 \\
        & Llama3-8B & 0.5 & 99.0 \\
    \midrule
    \multirow{5}{*}{max($\mu_{n_i} - 3\sigma_{n_i}$, $min_{n_i}$)}
        & GPT-2 & 0.5 & 100.0 \\
        & Pythia-1.4B & 0.1 & 96.5 \\
        & Mistral-7B-v0.1 & 11.1 & 99.0 \\
        & Phi-1.5 & 1.2 & 99.0 \\
        & Llama3-8B & 0.9 & 87.0 \\
    \midrule
    \multirow{5}{*}{min($\mu_{n_i} + 3\sigma_{n_i}$, $max_{n_i}$)}
        & GPT-2 & 7.8 & 99.0 \\
        & Pythia-1.4B & 1.5 & 100.0 \\
        & Mistral-7B-v0.1 & 2.3 & 84.0 \\
        & Phi-1.5 & 1.0 & 95.0 \\
        & Llama3-8B & 99.5 & 99.0 \\
    \midrule
    \multirow{5}{*}{$\text{Median}_{n_i}$}
        & GPT-2 & 0.2 & 99.0 \\
        & Pythia-1.4B & 0.1 & 74.5 \\
        & Mistral-7B-v0.1 & 0.5 & 92.0 \\
        & Phi-1.5 & 1.1 & 99.0 \\
        & Llama3-8B & 0.1 & 84.0 \\
    \midrule
    \multirow{5}{*}{$\text{Mode}_{n_i}$}
        & GPT-2 & 93.8 & 100.0 \\
        & Pythia-1.4B & 0.1 & 68.5 \\
        & Mistral-7B-v0.1 & 0.5 & 87.0 \\
        & Phi-1.5 & 1.3 & 98.0 \\
        & Llama3-8B & 0.1 & 60.5 \\
    \bottomrule
    \end{tabular}%
    }
    \caption{Ablation value-wise Global Transition Scores (\%) for entropy neurons ablation. The ablation values are computed over the knowledge probing dataset for each neuron activation distribution $n_i$ as illustrated in Figure \ref{fig:neurons_distro}. Specifically they consist of: the mean $\mu_{n_i}$, the mode Mode$_{n_i}$, the median Median$_{n_i}$, and two extreme values min($\mu_{n_i} + 3\sigma_{n_i}$, $max_{n_i}$), max($\mu_{n_i} - 3\sigma_{n_i}$, $min_{n_i}$) where $\sigma_{n_i}$ is the standard deviation. For the extreme values, we make sure to take the min$_{n_i}$/max$_{n_i}$ when $\mu_{n_i} \pm 3\sigma_{n_i}$ is out of distribution.}
    \label{tab:global_transition_ratio_by_ablation_value}
\end{table*}
%
%
\begin{figure*}[t]
    \centering
    \begin{subfigure}[b]{0.49\textwidth}
        \centering
        \includegraphics[width=\textwidth]{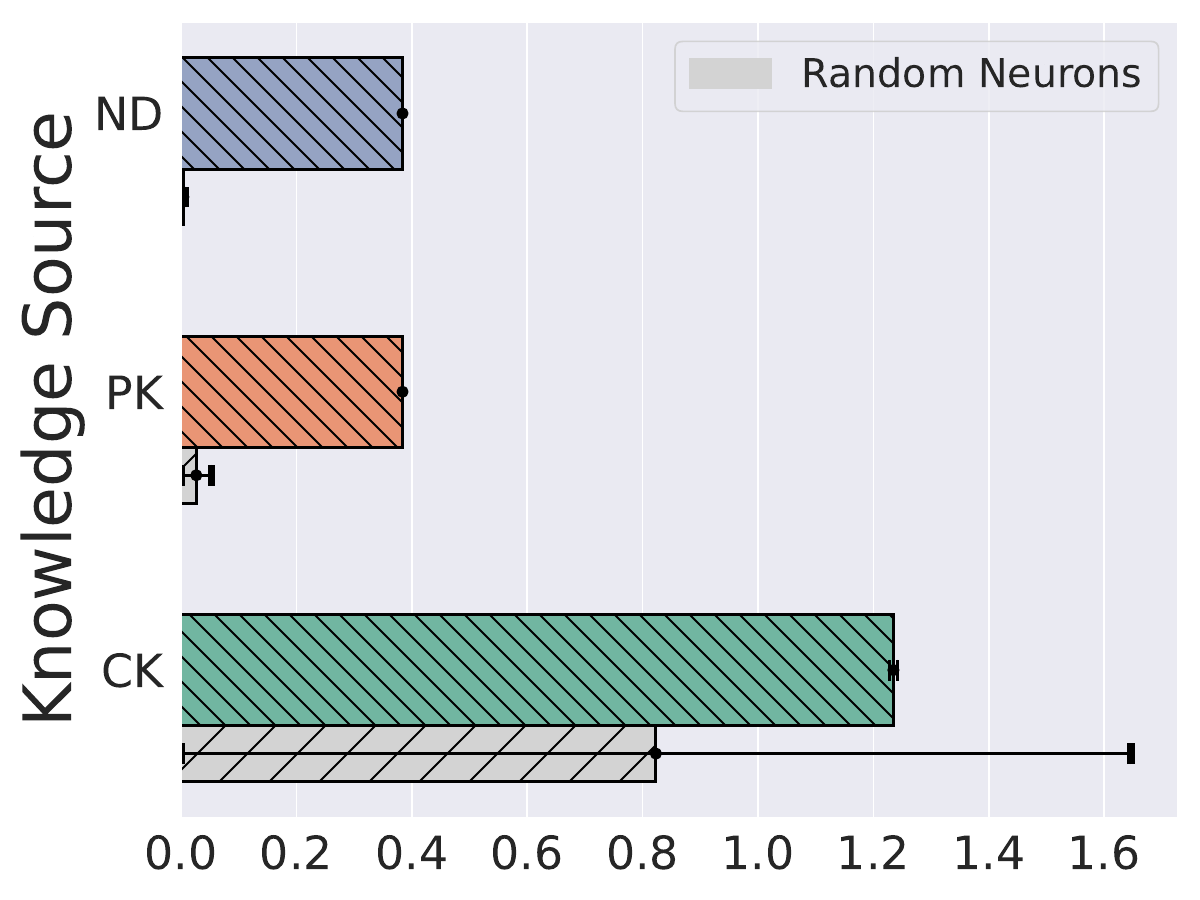}
        \caption{Llama-3-8B}
    \end{subfigure}
    \begin{subfigure}[b]{0.49\textwidth}
        \centering
        \includegraphics[width=\textwidth]{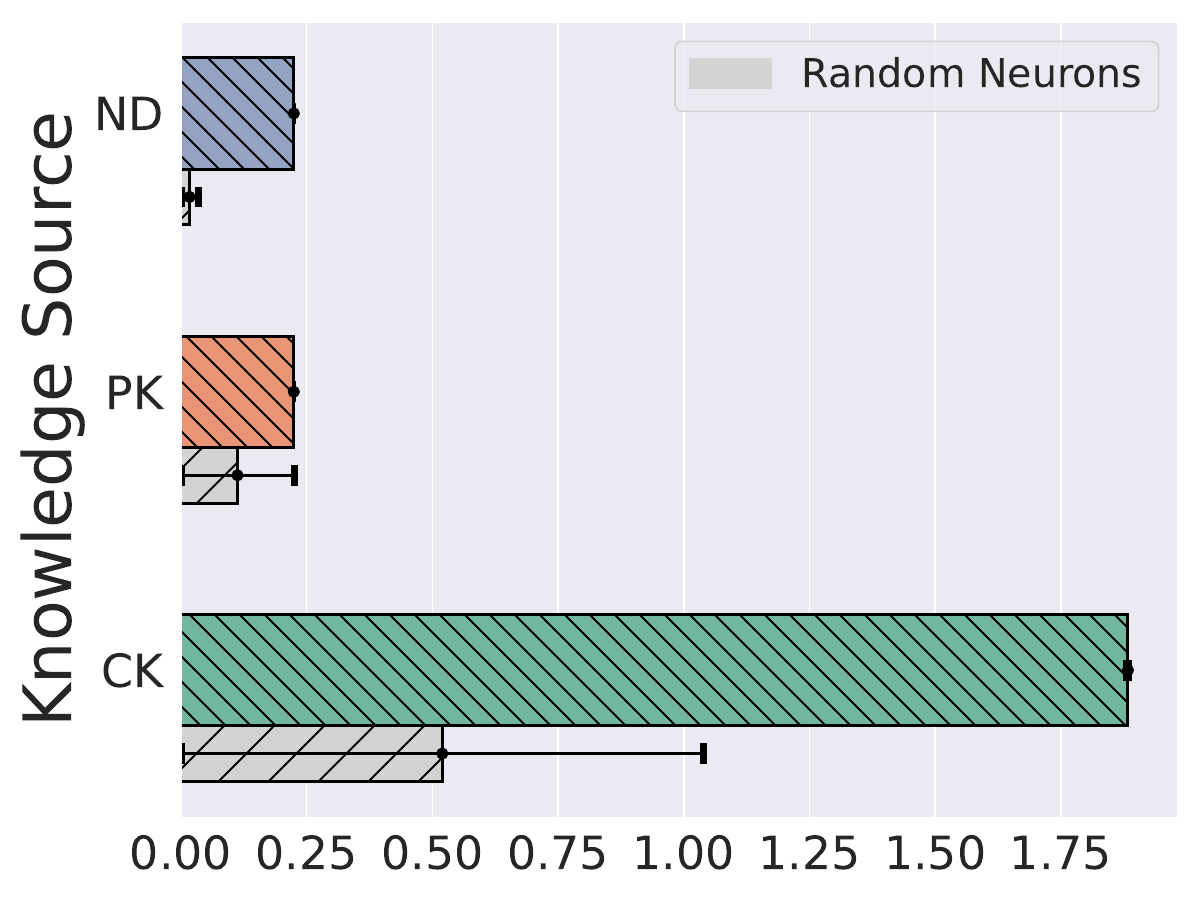}
        \caption{Mistral-7B}
    \end{subfigure}\\
    \begin{subfigure}[b]{0.49\textwidth}
        \centering
        \includegraphics[width=\textwidth]{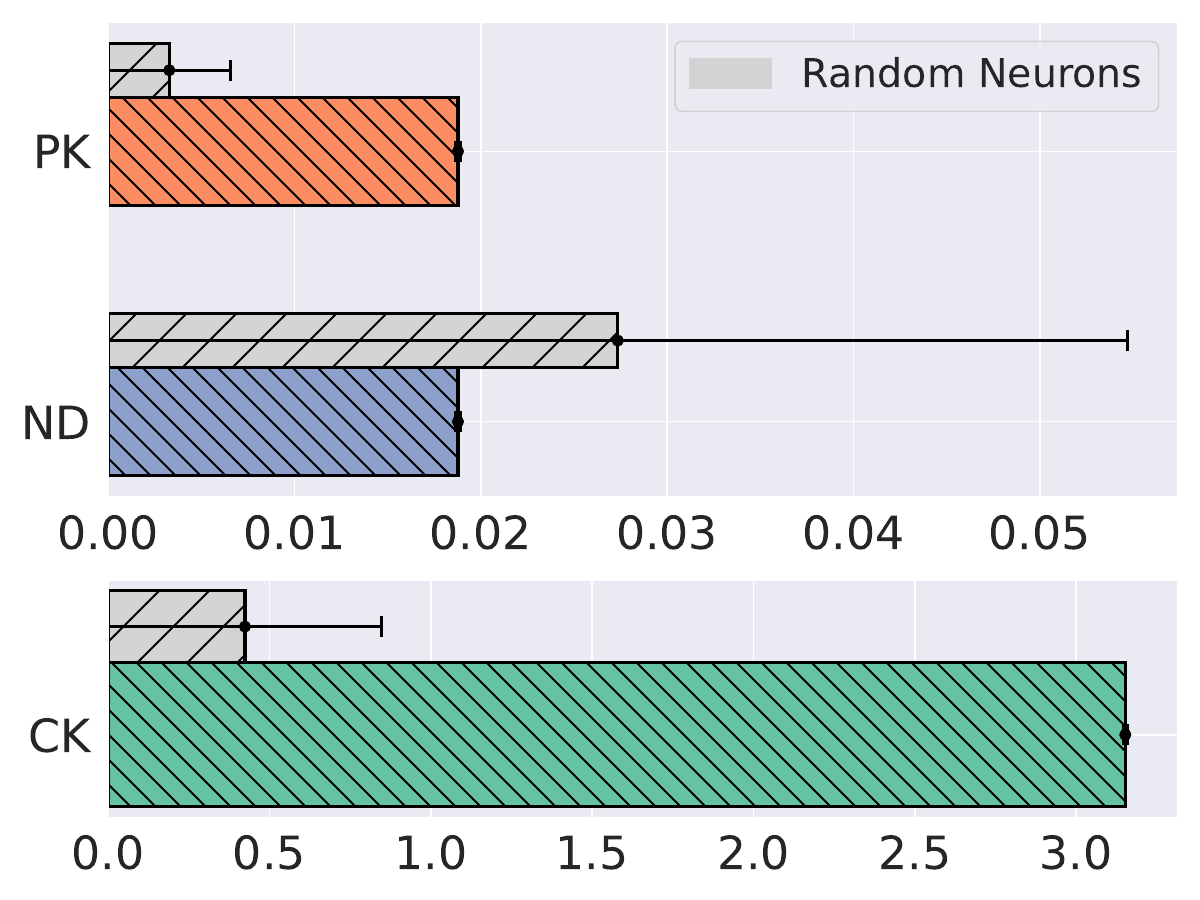}
        \caption{GPT2}
    \end{subfigure}
    \begin{subfigure}[b]{0.49\textwidth}
        \centering
        \includegraphics[width=\textwidth]{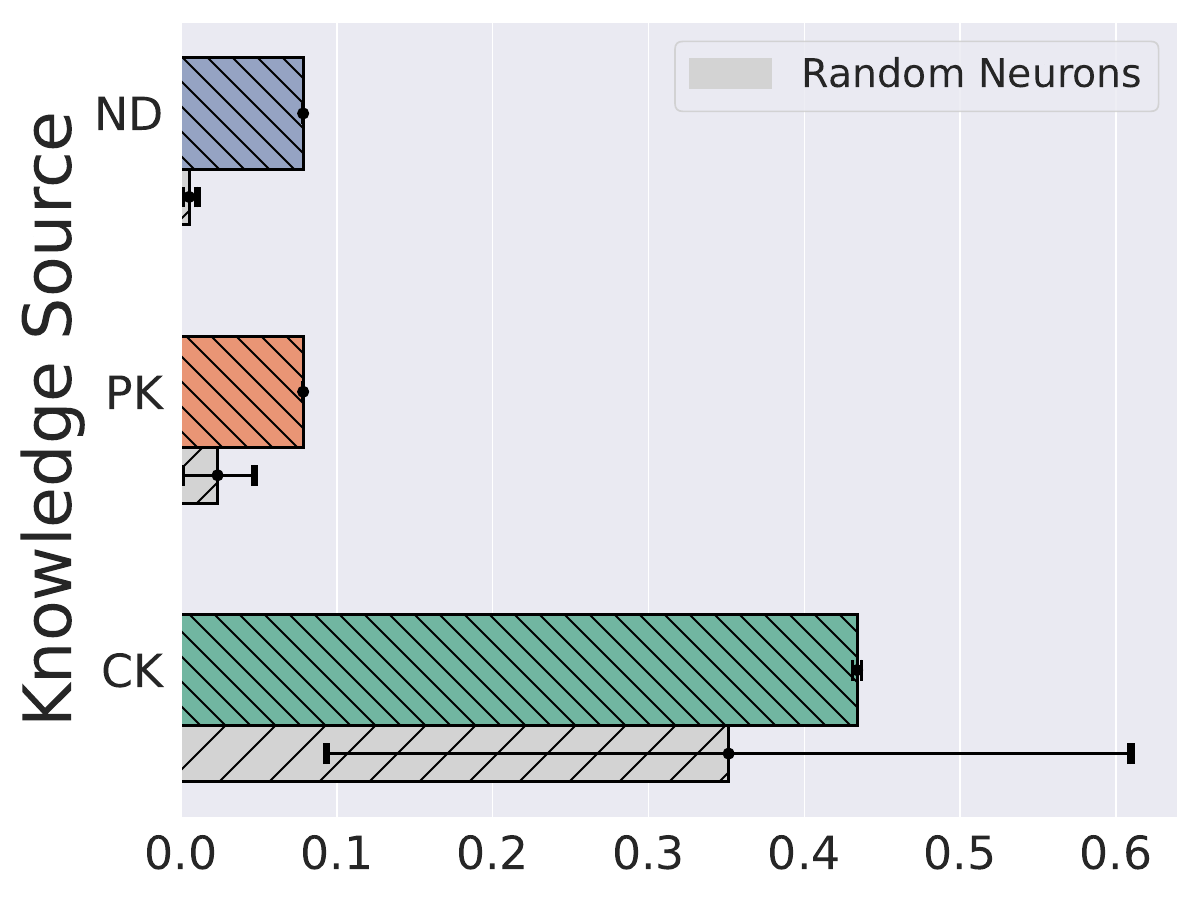}
        \caption{Pythia-1.4B}
    \end{subfigure}
    \caption{\textbf{Conversion Ratio (\%)}}
    \label{fig:conversion_ratio_llama_etal}
\end{figure*}
\begin{figure*}[t]
    \centering
    \begin{subfigure}[b]{0.32\textwidth}
        \centering
        \includegraphics[width=\textwidth]{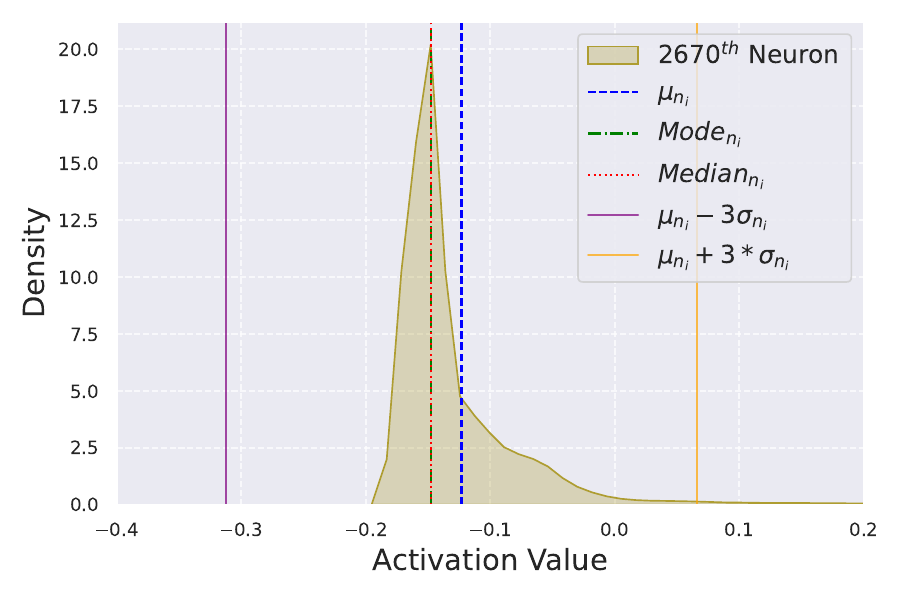}
        \caption{GPT2 (GELU)}
    \end{subfigure}
    \begin{subfigure}[b]{0.32\textwidth}
        \centering
        \includegraphics[width=\textwidth]{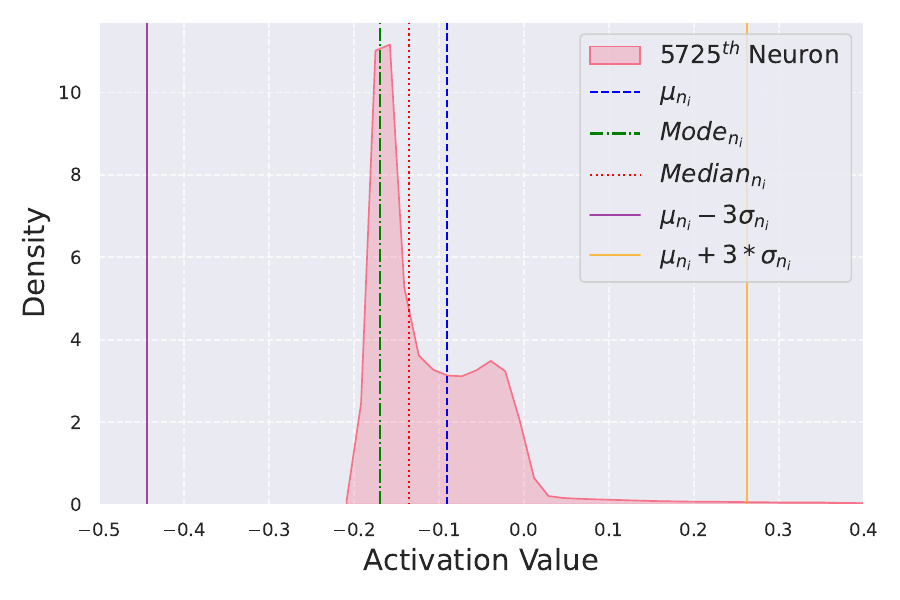}
        \caption{Phi-1.5 (GLU)}
    \end{subfigure}
    \begin{subfigure}[b]{0.32\textwidth}
        \centering
        \includegraphics[width=\textwidth]{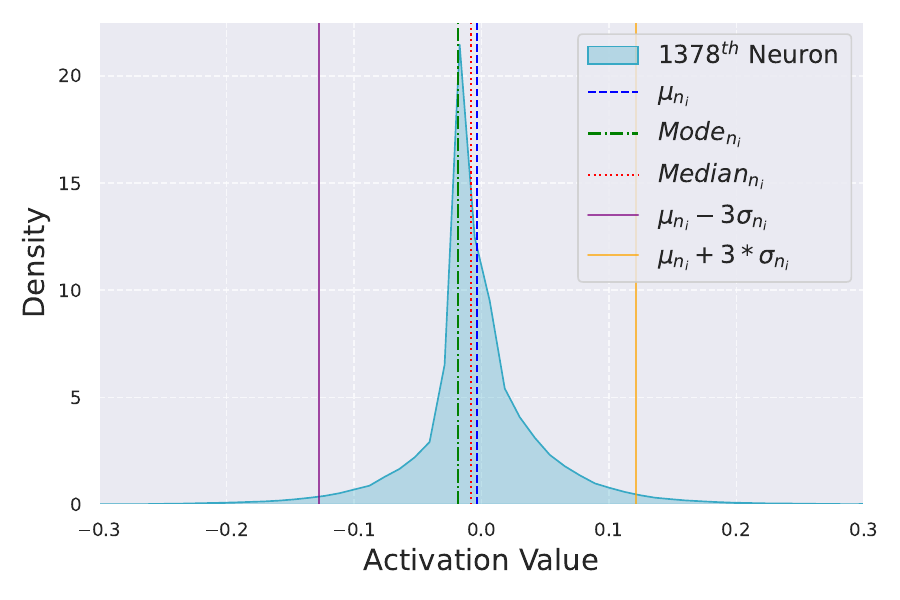}
        \caption{Llama-3-8B (SwiGLU)}
    \end{subfigure}
    \caption{Example of neurons distribution for each model as well as the ablation values. The Neuron where randomly selected for each model and the distribution was estimated based on the knowledge probing dataset \cite{tighidet2024probinglanguagemodelsknowledge}.}
    \label{fig:neurons_distro}
\end{figure*}
%
%
\begin{figure*}[t]
    \centering
    \begin{subfigure}[b]{0.49\textwidth}
        \centering
        \includegraphics[width=\textwidth]{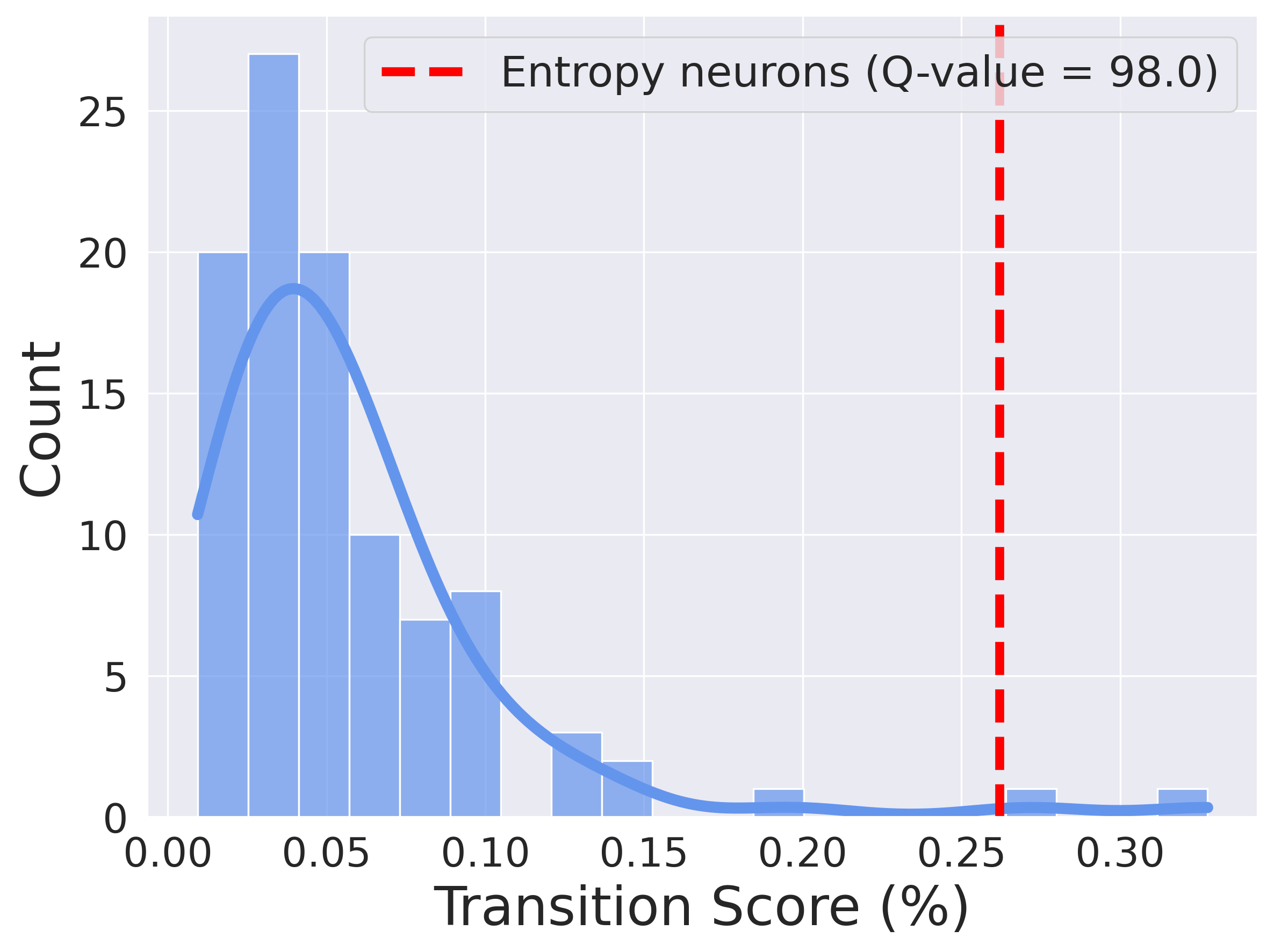}
        \caption{GPT2}
    \end{subfigure}
    \begin{subfigure}[b]{0.49\textwidth}
        \centering
        \includegraphics[width=\textwidth]{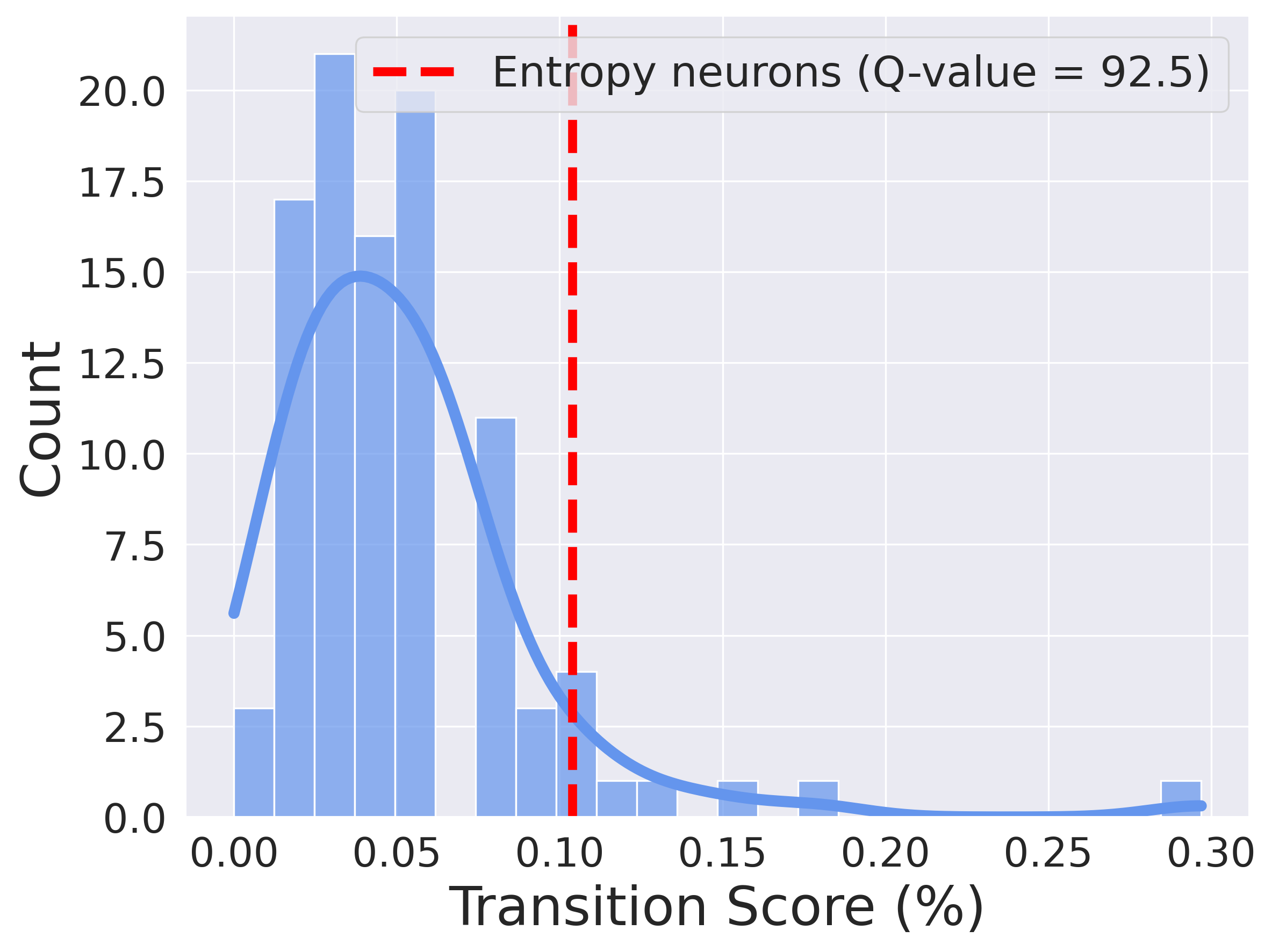}
        \caption{Pythia-1.4B}
    \end{subfigure}\\
    \begin{subfigure}[b]{0.49\textwidth}
        \centering
        \includegraphics[width=\textwidth]{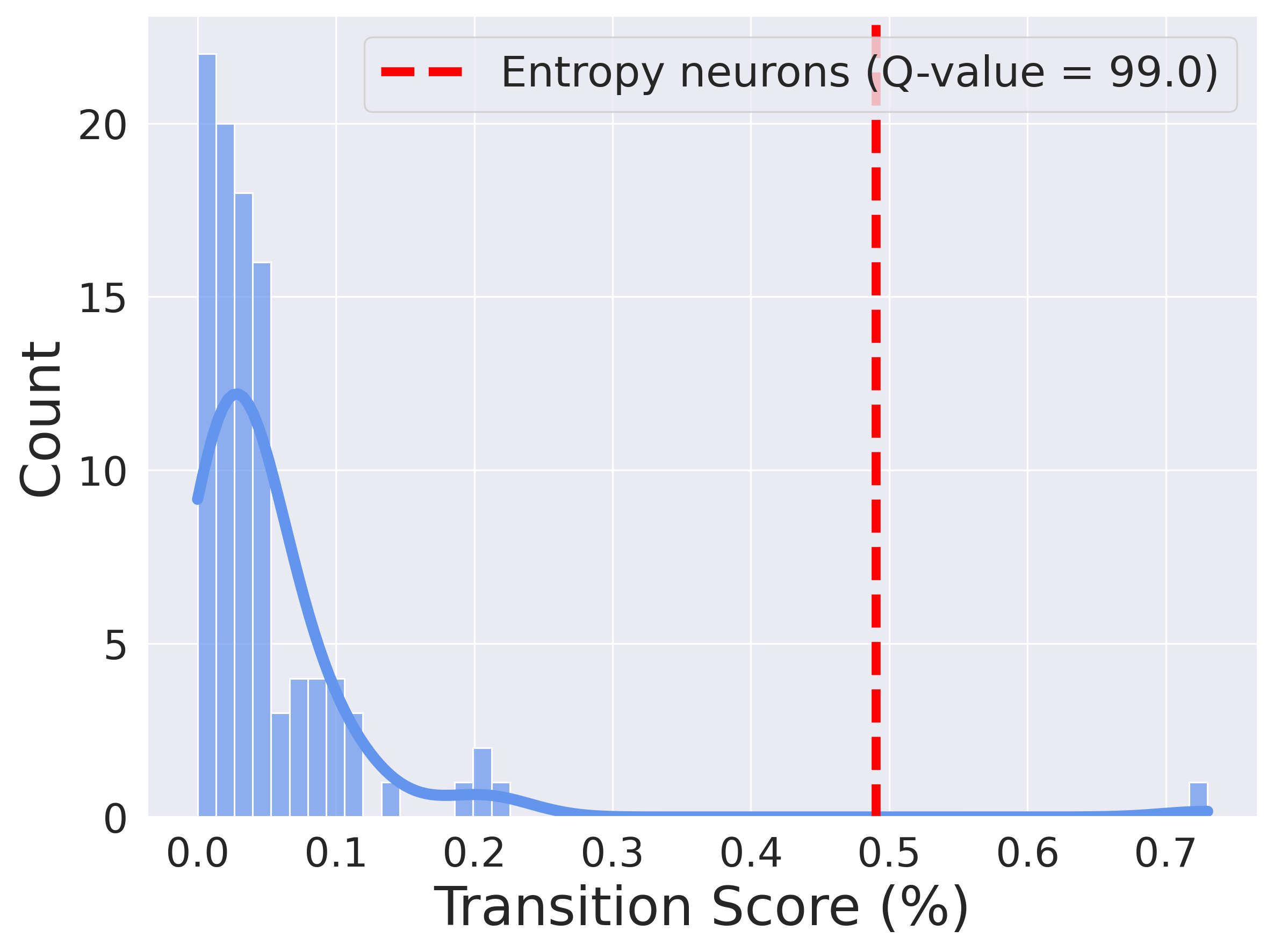}
        \caption{Llama-3-8B}
    \end{subfigure}
        \begin{subfigure}[b]{0.49\textwidth}
        \centering
        \includegraphics[width=\textwidth]{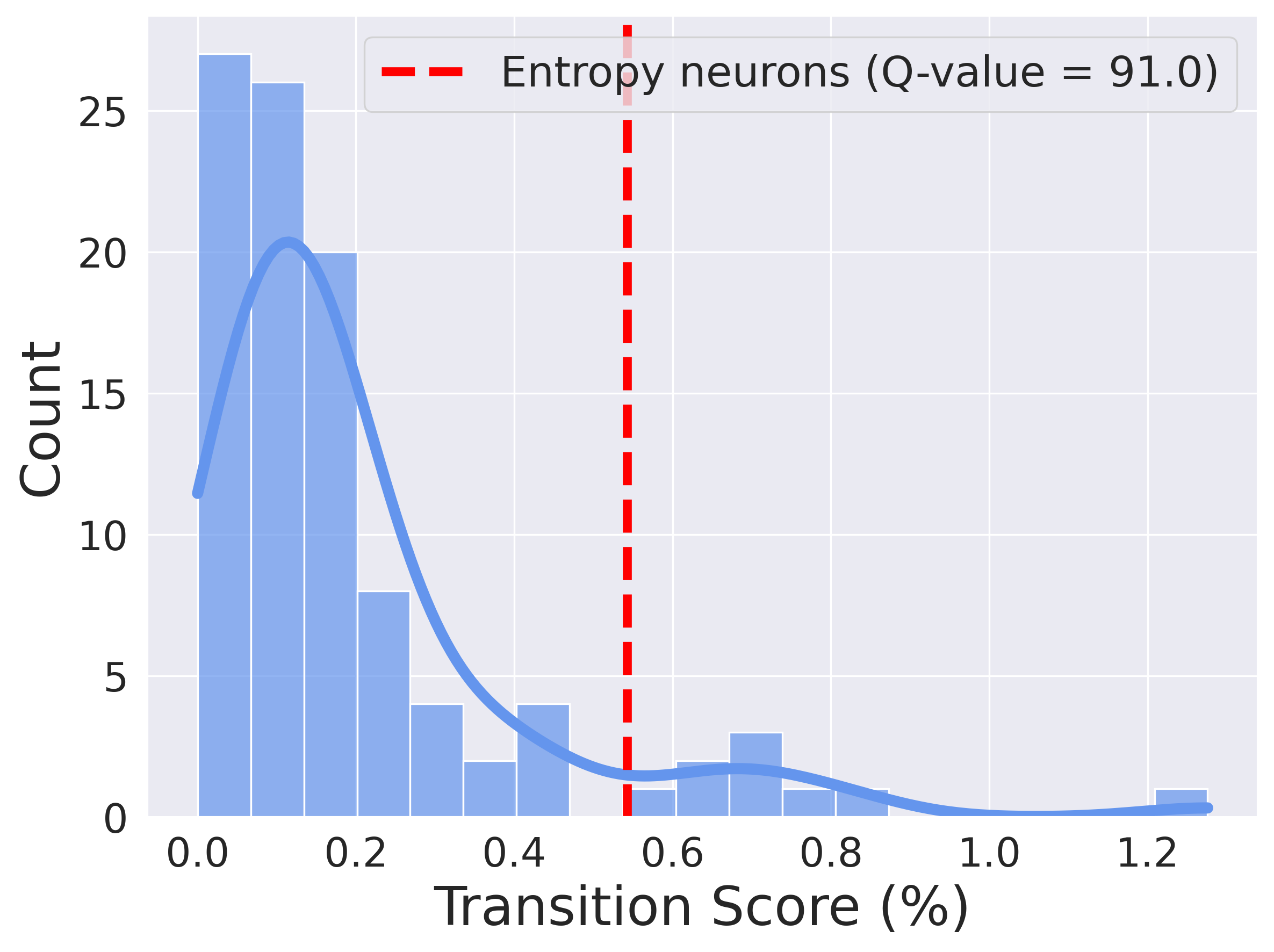}
        \caption{Mistral-7B-v0.1}
    \end{subfigure}
    \caption{Global Transition Scores, ablating entropy neurons exhibit a higher transition in the used knowledge sources compared to 100 sets of random neurons which indicates the unique property of entropy neurons to affect the knowledge source to select.}
    \label{fig:gts_rest_of_models}
\end{figure*}

\end{document}